\documentclass[lettersize,journal]{IEEEtran}
\usepackage{cite}
\usepackage{moreverb,url}
\usepackage{booktabs}
\usepackage{siunitx}
\usepackage{multirow}
\usepackage{pifont}
\usepackage{amssymb}
\usepackage{amsmath}
\usepackage{subfigure}
\usepackage{tabularx}
\usepackage{comment}
\usepackage{mathtools} 
\usepackage{stmaryrd}
\usepackage{tabularx} 
\usepackage{color}
\usepackage{colortbl}
\usepackage[dvipsnames]{xcolor}
\usepackage{caption}
\usepackage[dvipsnames]{xcolor}
\usepackage{pifont}       
\usepackage{bbding}
\usepackage{booktabs}
\usepackage[colorlinks,bookmarksopen,bookmarksnumbered,citecolor=red,urlcolor=red]{hyperref}
\usepackage{colortbl} 
\usepackage{arydshln}
\usepackage[dvipsnames]{xcolor}
\usepackage{pifont} 
\usepackage{tikz} 
\newcommand\BibTeX{{\rmfamily B\kern-.05em \textsc{i\kern-.025em b}\kern-.08em
T\kern-.1667em\lower.7ex\hbox{E}\kern-.125emX}}

\definecolor{C1}{HTML}{ECA8A9}
\definecolor{C2}{HTML}{74AED4}
\definecolor{C3}{HTML}{D3E2B7}
\definecolor{C4}{HTML}{CFAFD4}
\definecolor{C5}{HTML}{F7C97E}
\definecolor{C6}{HTML}{E8D5C4}
\definecolor{C7}{HTML}{EEEEEE}
\definecolor{C8}{HTML}{BCEE68}

\newcommand{\cmark}{\ding{51}}%
\newcommand{\xmark}{\ding{55}}%

\definecolor{tabfirst}{rgb}{1, 0.75, 0.7}
\definecolor{tabsecond}{rgb}{1, 0.85, 0.65}
\definecolor{tabthird}{rgb}{1, 0.96, 0.7}
\definecolor{cvprblue}{rgb}{0.21,0.49,0.74}

\begin{document}

\title{MCN-SLAM: Multi-Agent Collaborative Neural SLAM with Hybrid Implicit Neural Scene Representation }

\author{Tianchen Deng, Guole Shen, Xun Chen, Shenghai Yuan, Hongming Shen, Guohao Peng, Zhenyu Wu, Jingchuan Wang, Lihua Xie, Fellow, IEEE,  Danwei Wang, Fellow, IEEE, Hesheng Wang, Weidong Chen}

\twocolumn[{%
\renewcommand\twocolumn[1][]{#1}%

\maketitle

\begin{center}
  \centering
  \vspace{-1cm}
  \captionsetup{type=figure}
  \includegraphics[width=\linewidth]{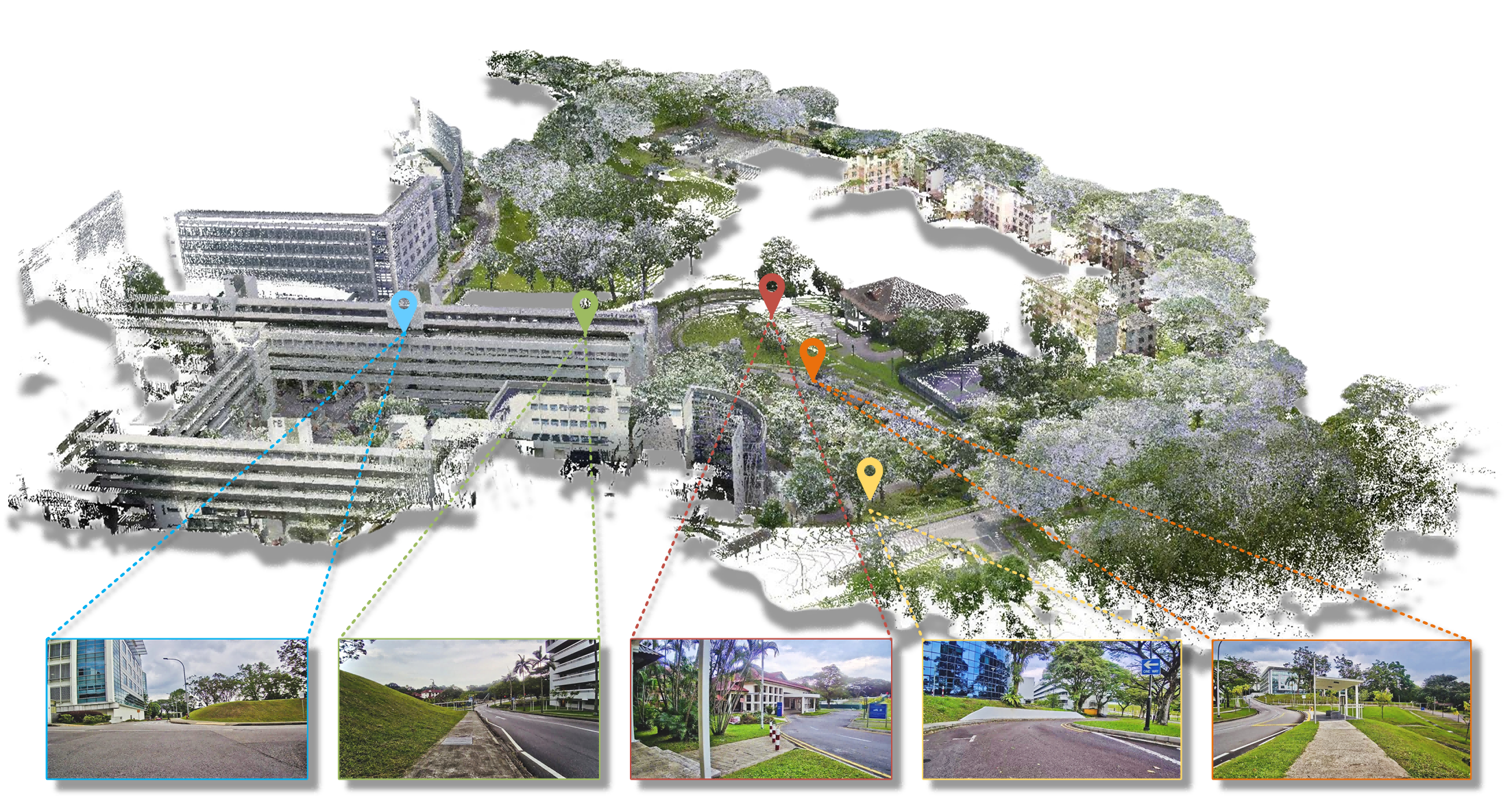}
  \setlength{\abovecaptionskip}{-5pt}
  \captionof{figure}{We visualize the outdoor scenes of our DES dataset ($\approx $276,000 $m^2$). The 3D map collected by more than 200 industrial laser scanners. We show the scene reconstruction results of some localized regions in different colors.  }
  \label{fig:teaser}
    
\end{center}%
}]

\makeatletter{\renewcommand*{\@makefnmark}{}
 \footnotetext{Tianchen Deng, Guole Shen, Hesheng Wang, Jingchuan Wang, Weidong Chen are with Institute of Medical Robotics and Department of Automation, Shanghai Jiao Tong University, and Key Laboratory of System Control and Information Processing, Ministry of Education, Shang hai 200240, China. Xun Chen, Shenghai Yuan, Hongming Shen, Guohao Peng, Zhenyu Wu, Lihua Xie, Danwei Wang is with School of Electrical and Electronic Engineering, Nanyang Technological University, Singapore. This research is supported by the National Research Foundation, Singapore, under the NRF Medium Sized Centre scheme (CARTIN), ASTAR under National Robotics Programme with Grant No.M22NBK0109, and by the National Research Foundation, Singapore and Maritime and Port Authority of Singapore under its Maritime Transformation Programme (Project No. SMI-2022-MTP-04). This work is supported by the National Key R\&D Program of China (Grant 2020YFC2007500), the National Natural Science Foundation of China (Grant U1813206), and the Science and Technology Commission of Shanghai Municipality (Grant 20DZ2220400). This research is supported by the National Research Foundation, Singapore, under the NRF Medium Sized Centre scheme (CARTIN), Maritime and Port Authority of Singapore under its Maritime Transformation Programme (Project No. SMI-2022-MTP-04), ASTAR under National Robotics Programme with Grant No.M22NBK0109. }

\markboth{IEEE Transactions on Pattern Analysis and Machine Intelligence}%
{Shell \MakeLowercase{\textit{et al.}}: A Sample Article Using IEEEtran.cls for IEEE Journals}

\IEEEpubidadjcol

\begin{abstract}

Neural implicit scene representations have recently shown promising results in dense visual SLAM. However, existing implicit SLAM algorithms are constrained to single-agent scenarios, and fall difficulty in large-scale scenes and long sequences. Existing NeRF-based multi-agent SLAM frameworks cannot meet the constraints of communication bandwidth. To this end, we propose the first distributed multi-agent collaborative neural SLAM framework with hybrid scene representation, distributed camera tracking, intra-to-inter loop closure, and online distillation for multiple submap fusion. A novel triplane-grid joint scene representation method is proposed to improve scene reconstruction. Moreover, for large-scale and unbounded outdoor scene, we propose a novel normalization method for implicit scene representation, which enhances the network's capability for extensive unbounded environments. A novel intra-to-inter loop closure method is designed to achieve local (single-agent) and global (multi-agent) consistency. We also design a novel online distillation method to fuse the information of different submaps to achieve global consistency. Furthermore, to the best of our knowledge, there is no real-world dataset for NeRF-based/GS-based SLAM that provides both continuous-time trajectories groundtruth and high-accuracy 3D meshes groundtruth. To this end, we propose the first real-world Dense slam (DES) dataset covering both single-agent and multi-agent scenarios, ranging from small room to large-scale outdoor scenes, with high-accuracy ground truth for both 3D mesh and continuous-time camera trajectory. This dataset can advance the development of the research in both SLAM, 3D reconstruction, and visual foundation model. Experiments on
various datasets demonstrate the superiority of the proposed method in both mapping, tracking, and communication. The dataset and code will open-source on \href{https://github.com/dtc111111/mcnslam}{https://github.com/dtc111111/mcnslam}.

\end{abstract}

\section{Introduction}
\label{sec:intro}

Visual Simultaneous Localization and Mapping has been a fundamental challenge in robotics and computer vision, which aims to achieve 3D reconstruction of unknown environments and localization. Visual SLAM has wide applications such as Spatial AI~\cite{foundation}, autonomous driving~\cite{prosgnerf}, and remote sensing~\cite{magmm}. Traditional visual SLAM has witnessed continuous development, achieving accurate mapping and tracking in various scenarios, such as ~\cite{orbslam,vins,shenslam}. They use sparse point cloud as the scene representation which use handcraft descriptors for image matching and represent scenes using sparse feature point maps. Due to the sparse nature of such point cloud, it is difficult for humans to understand how machines interact with the scene, and these methods cannot meet the demands of collision avoidance and motion planning.  Attention then turns to dense scene reconstruction. DTAM~\cite{dtam} is the first dense visual SLAM system, which estimates detailed textured depth maps at selected keyframes to produce a surface patchwork with millions of vertices. \cite{kintinuous, elasticfusion, kinectfusion} are the subsequent improvement works focusing on scene reconstruction accuracy. They reconstruct meaningful 3D global maps with dense scene representation. However, their performance is limited by high memory consumption, slow processing speeds, and low accuracy in real-time operation. Nowadays, with the proposal of Neural Radiance Fields (NeRF)~\cite{NeRF}, many following works combine implicit scene representation with SLAM framework. iMAP~\cite{imap} is the first work to use a single multi-layer perceptron (MLP) to reconstruct the scene in an online mapping framework. \cite{niceslam,eslam,coslam,plgslam} further improve scene representation. 
However, most existing NeRF-based SLAM methods focus on single agent settings.  In contrast, multi-robot collaborative simultaneous localization and mapping (SLAM) represents a crucial area of robotics research, owing to its ability to enable situational awareness across large-scale environments over extended durations. This capability is foundational for a wide range of applications, including factory automation, search and rescue operations, intelligent transportation systems, planetary exploration, as well as surveillance and monitoring in both military and civilian contexts. Traditional multi-agent SLAM systems~\cite{multiagent1,multiagent2} also use sparse point cloud as the scene representation. Some methods employs dense semantic segmentation networks~\cite{salt,sfpnet} for multi-agent semantic mapping and pose estimation. Kimera-multi~\cite{kimera-multi} propose a complete distributed multi-robot system for semantic dense visual SLAM, which is capable of operating under realistic communication constraints.    
CP-SLAM~\cite{CP-SLAM} is the first multi-agent SLAM neural system with implicit scene representation using a centralized learning framework for scene reconstruction. By integrating NeRF, this approach significantly enhances dense scene reconstruction. However, it operates as a centralized system, requiring all keyframe images to be transmitted to a central server to train an additional global network, which cannot meet the communication bandwidth restriction. Given the constraints of privacy and communication bandwidth, we design the first distributed multi-agent neural SLAM framework, which only relies on local (peer-to-peer) communication with only neural network parameters,
keyframe descriptors instead of raw data when the system detects a loop. Compared to point cloud map representations, our neural scene representation method does not grow indefinitely as the map size increases, making it more efficient for multi-agent communication and data transmission. Our method achieves accurate scene reconstruction, robust pose estimation, and efficient multi-agent communication.   The system can continue to operate even if communication is lost with a particular agent. 
\begin{figure*}[h]
    \centering
    \includegraphics[width=\linewidth]{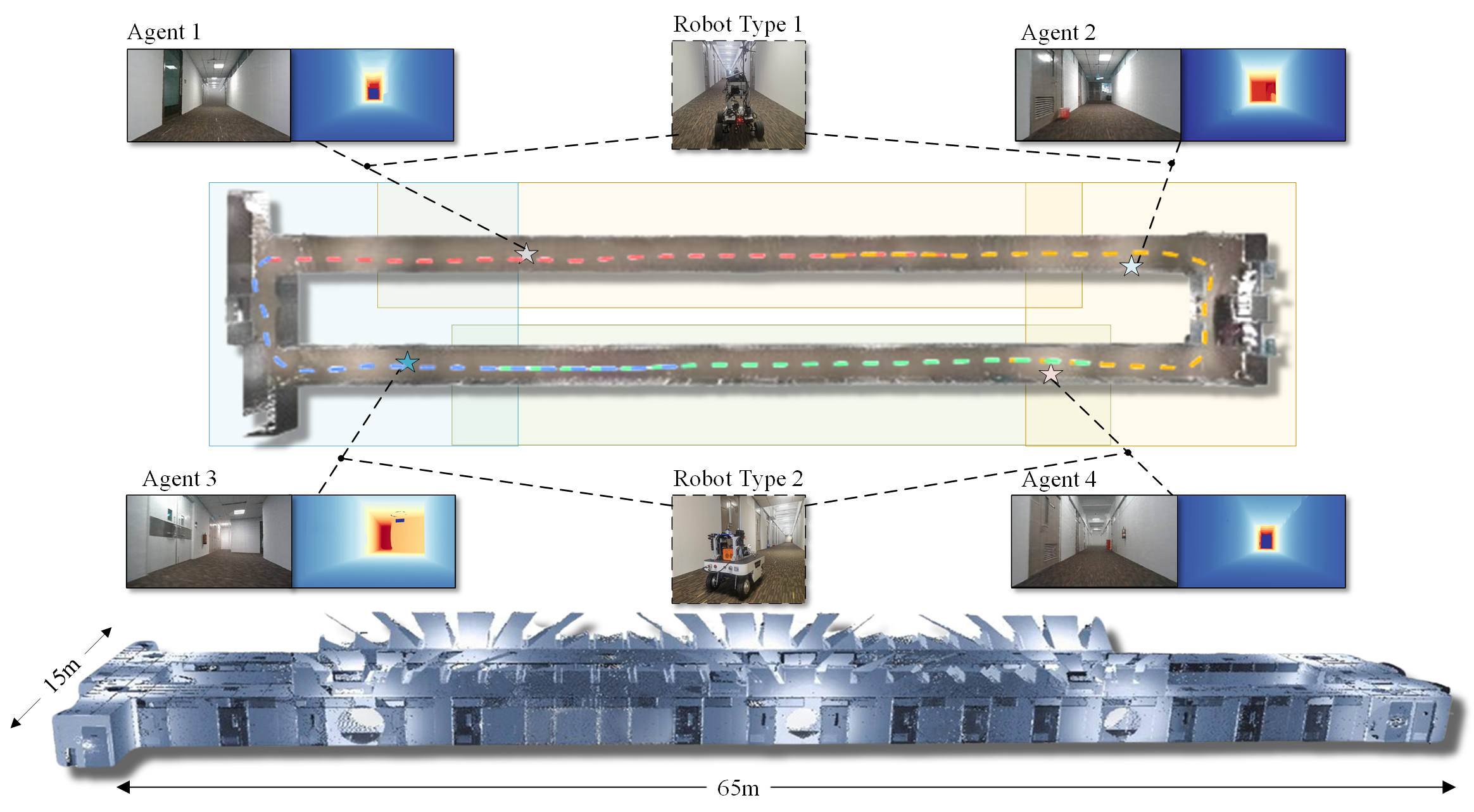}
    
    \caption{We present MCN-SLAM, the first distributed multi-agent collaborative SLAM system with distributed mapping and camera tracking, hybrid implicit scene representation, intra-to-inter loop closure, and multiple submap fusion.
  Depicted at the middle, we present the scene reconstruction results of four agents, demonstrating scene reconstruction performance in the real-world, large-scale long-corridor scenes ($\approx $1200 $m^2$). This scene is collected through various industrial laser scanners.  We present the rendered depth and color image of different type of agents around the corridor. The trajectory of each agent is marked in a unique color for clarity.   }
     
    \label{fig:teaser2}
\end{figure*}

In local scene representation, we propose a hybrid scene representation method with planar-grid-coordinate encoding network architecture for fast convergence, high accuracy, and completion of unseen region. 
We utilize low-resolution tri-plane feature to capture the scene’s low-frequency components (e.g. shape), which facilitates fast convergence. Meanwhile, high-resolution hash grid are employed to represent high-frequency details (e.g. texture), enabling high-fidelity reconstruction.
We use one-blob encoding (coordinate encoding) with Multilayer Perceptron (MLP) to represent the coherence priors inherent for hole filling. 
 The novel hybrid scene representation method brings together the benefits of  fast convergence, high accuracy, and hole filling in areas without observation. Specifically, for real-world outdoor unbounded scenes, we design a normalized representation method to improve the network capability.

Furthermore, we propose a novel intra-to-inter loop closure method with online distillation for the multi-agent neural SLAM framework to mitigate the accumulation of pose errors and achieve consensus scene reconstruction among multiple agents.
Intra-loop closure is designed to detect previously visited locations by a single robot, effectively eliminating cumulative errors using the current robot’s keyframe database.
In a distributed neural SLAM system, all these agents optimize their scene representation independently with local and different observations. As a result, the final maps generated by individual agents only represent specific portions of the scene, leading to discrepancies.
So, we propose a novel inter-loop closure to detect the same scenario visited by multiple agents, register submaps generated from different agents, and fuse the different implicit submaps. A global consistency loss is introduced to optimize and register the pose of different submaps. A peer-to-peer online distillation method is designed for the fusion of different neural implicit submaps. When communication is available, we perform intra-to-inter loop closure, exchange the neural network parameters of each agent, and perform implicit submap fusion to achieve global consistency across multiple agents. The results demonstrate that each agent ultimately acquires accurate global information, enabling the construction of a consistent global map.

Furthermore, to the best of our knowledge, we find that current datasets are either virtual, such as Replica~\cite{replica} or only provide trajectory ground truth without 3D ground truth, such as ScanNet~\cite{scannet}, and TUM dataset~\cite{tum}. ScanNet++~\cite{scannetpp} is a recently proposed dataset that offers 3D ground truth. However, its poses are obtained through COLMAP, making them unsuitable as accurate SLAM ground truth. In addition, Scannet++ is primarily intended as a NeRF Training $\&$ Novel View Synthesis dataset, as stated in their paper and SplaTAM~\cite{splatam}. ScanNet++ contains non-time-continuous trajectories with numerous abrupt jumps and teleportations, which makes it unsuitable for SLAM systems. To this end, we propose a real-world dataset ranging from small-room scenarios to large-scale campus. Our dataset provides high-accuracy and time-continuous trajectory and 3D mesh groundtruth, which is suitable for all neural SLAM and traditional SLAM systems, such as NeRF/3DGS-based SLAM systems and 3D reconstruction methods such as DUSt3R~\cite{dust3r} and VGGT~\cite{vggt}.
Overall, our contributions are shown as follows:
\begin{itemize}
\item We design the first distributed multi-agent neural SLAM framework, MCN-SLAM, with hybrid neural scene representation, distributed mapping
and tracking, intra-to-inter loop closure, achieving accurate collaborative scene reconstruction, robust camera tracking, and efficient multi-agent communication.
\item A novel hybrid representation method with planar-grid-coordinate network is designed for fast convergence, accurate and efficient neural scene representation, substantially reducing memory growth from cubic to quadratic. We also design a novel normalized representation to improve scene capability for large-scale unbounded outdoor scenes.
\item We propose a novel intra-to-inter loop closure method with an online distillation method to eliminate the cumulative pose drift and register multiple submaps. The online distillation method is proposed to fuse the information from different neural submaps to achieve global map consistency. 

\item We propose the first real-world large-scale neural dense SLAM dataset (DES) for all kinds of SLAM systems with high-accuracy ground-truth for both camera trajectory and 3D reconstruction mesh, which can significantly foster the development of the community.
\end{itemize}

\begin{figure*}[h]
    \centering
    \includegraphics[width=\linewidth]{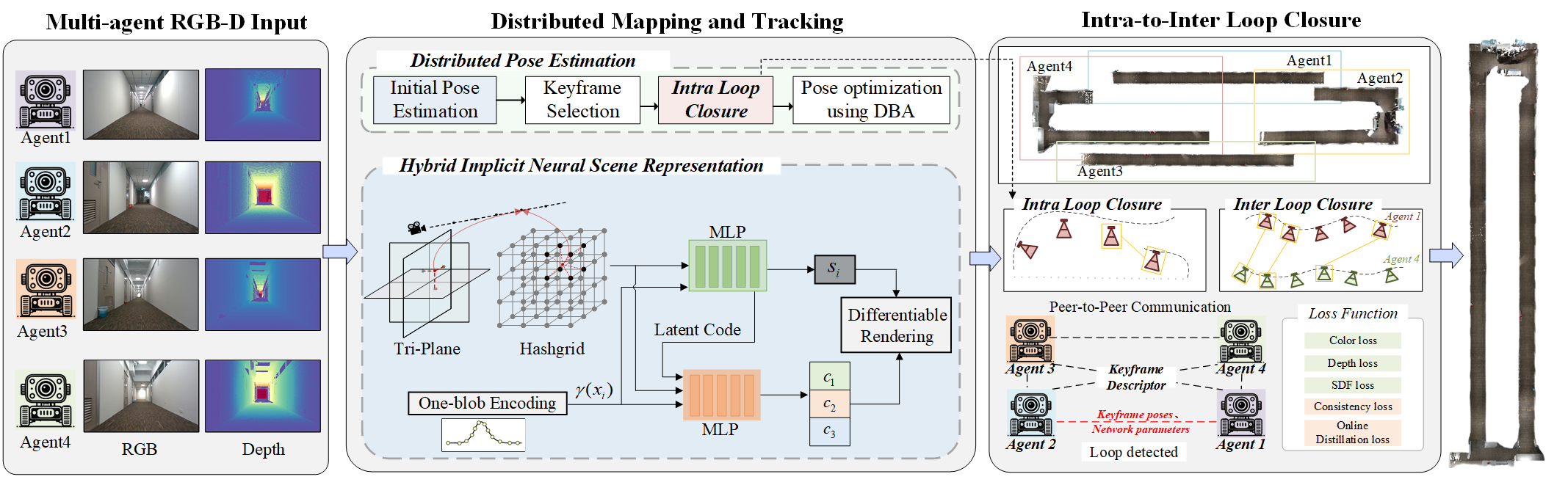}
     
    \caption{\textbf{System Overview.} Our system is a multi-agent collaborative SLAM system which consists hybrid scene representation, distributed tracking, intra-to-inter loop closure, and submap-fusion. In distributed optimization module, each agent takes the color images and depth images as input. In addition, each agent will exchange the network weights of its peers. We carefully design consistency loss with color, depth, and SDF loss in inter-loop closure. Each agent can successively performs individual scene mapping and collaborative mapping and tracking to generate the final neural implicit map with submap-fusion. }
    \label{fig:system}
     
\end{figure*}

\section{Related work}

\noindent \textbf{Sparse Visual SLAM}
In recent years, sparse visual SLAM has enhanced real-time performance by focusing on extracting and mapping key environmental features. PTAM~\cite{ptam} is a keyframe-based approach, is notable for separating feature tracking and map construction into two parallel tasks. ORB-SLAM~\cite{orbslam} is a milestone in sparse visual SLAM with its comprehensive design and multi-threaded architecture. Building upon the robust base of ORB-SLAM, ORB-SLAM2~\cite{orbslam2} expands capabilities to stereo and RGB-D cameras, enhancing accuracy and robustness across diverse environments. Further advancements include semi-direct approaches like SVO~\cite{svo}, which combine direct tracking with feature-based mapping. VINS-Mono~\cite{vins} leverages both monocular visual data and IMU inputs through tightly-coupled non-linear optimization, and Fast-LIVO2~\cite{fastlivo2} leverages lidar, camera and IMU inputs for pose estimation.   \cite{deng,xie,xie2} propose a novel feature selection method for visual SLAM system in dynamic scenes. However, the inherent limitation of sparse methods is their inability to provide detailed and continuous environmental representation. Sparse SLAM captures only key points, resulting in maps that lack comprehensive surface details, which is insufficient for tasks requiring detailed environmental interaction.

\noindent \textbf{Dense Visual SLAM}
Dense visual SLAM technologies have significantly advanced in the past two decades. DTAM~\cite{dtam} first introduces a dense SLAM system that leverages photometric consistency for tracking handheld cameras and depicting scenes as cost volumes. KinectFusion~\cite{kinectfusion} employs the iterative-closest-point method for camera tracking and updates the scene using TSDF-Fusion. DROID-SLAM~\cite{droidslam} leverages neural networks to extract richer contextual information from images, significantly enhancing trajectory estimation. Kimera~\cite{kimera} propose a 3D dynamic scene graph (DSG), that seamlessly captures metric and semantic aspects of a dynamic environment and estimate an accurate 3D metric–semantic mesh model in real-time.  DPVO~\cite{dpvo} introduces a novel recurrent update operator for patch based correspondence coupled with differentiable bundle adjustment, outperforming all prior work. However, these methods remain reliant on traditional SLAM scene representation (point clouds), which inherently limits their capabilities in complex scene reconstruction and require  substantial memory.

\noindent \textbf{Implicit Neural SLAM}
Recent advancements in neural implicit networks have revolutionized 3D scene reconstruction and camera pose estimation. Many efforts have focused on combining NeRF with SLAM systems to enhance both accuracy and efficiency in dynamic environments. iMAP introduces a real-time update framework that leverages a single MLP decoder for scene reconstruction. NICE-SLAM~\cite{niceslam} extends this by incorporating a learnable hierarchical feature grid, which improves the level of detail and allows for larger scene representations. ESLAM~\cite{eslam} further optimizes memory usage by using axis-aligned feature planes, while Co-SLAM~\cite{coslam} adopts a joint coordinate-parametric encoding strategy to ensure complete and accurate surface reconstruction, including unobserved areas. To adapt to larger environments, \cite{plgslam,mipsfusion,ngmslam} enhances scalability with its progressive scene representation method. Point-SLAM~\cite{pointslam} and Pin-SLAM~\cite{pinslam} uses dense neural points and corresponding MLP to represent the scene. GO-SLAM~\cite{goslam}, Loopy-SLAM~\cite{loopyslam} use loop closure method to enhance camera tracking performance.
NeSLAM~\cite{neslam} uses depth uncertainty and design a depth completion and denoising network for sparse depth images input. DDN-SLAM~\cite{ddnslam} use neural implicit representation and dynamic masks for dynamic scenes.
SNI-SLAM~\cite{snislam} use semantic feature with neural scene representation to improve the scene reconstruction performance.
Some 3DGS-based methods, such as \cite{splatam,compact,sgsslam,mgslam,loopsplat,vpgsslam} use 3D Gaussian scene representation method for accurate and fast reconstruction. Our method builds on these approaches by expanding from single-agent to multi-agent systems, achieving a globally coherent 3D reconstruction in large-scale scenarios. 

\noindent \textbf{Collaborative Visual SLAM}
The goal of collaborative visual SLAM is to estimate the relative poses between multiple robots while creating a consistent global map. This process typically employs two primary architectures: centralized and distributed. 
The centralized architecture relies on a central server, where all agents communicate their observations and receive updates, enabling efficient data integration and global consistency. CVI-SLAM~\cite{CVI-SLAM} is the first to introduce full visual-inertial collaborative SLAM with two-way communication. CoSLAM~\cite{coslam2} studies the problem of vision-based simultaneous localization and mapping (SLAM) in dynamic environments. All cameras work together to build a global map, including 3D positions of static background points and trajectories of moving foreground points.  CCM-SLAM~\cite{ccmslam} proposes a centralized collaborative SLAM framework for robotic agents, each equipped with a monocular camera, a communication unit, and a small processing board. It optimizes server by offloading heavy computations while maintaining agent autonomy with onboard visual odometry. 
Some studies focus on decentralized collaborative SLAM, where agents exchange information directly with each other and perform all data fusion onboard, without relying on a central instance to coordinate the system. Kimera-Multi~\cite{kimera-multi}  is a multi-robot Visual-Inertial Odometry (VIO) system that enhances tracking accuracy using IMU data, distinguishing it from typical Visual Odometry (VO) systems. Door-SLAM~\cite{DOOR-SLAM} implements a distributed Pose-Graph Optimization (PGO) scheme.  Some works integrate implicit neural representations into multi-agent SLAM systems.  Swarm-SLAM~\cite{swarmslam} and $D^2$SLAM~\cite{D2slam} are scalable, flexible, decentralized, and sparse SLAM systems, which are all key properties in swarm robotics. Some methods~\cite{macim,CP-SLAM,mneslam} introduce a multi-agent collaborative implicit mapping and tracking algorithm to construct  maps. However, these approaches suffer from limitations such as inefficient inter-robot communication, lack of global map consistency, or poor adaptability to both indoor and outdoor environments. Our proposed multi-agent neural SLAM framework is the first distributed method, achieving high-accuracy and consistent scene reconstruction, efficient robot communication, and robust adaptability across both indoor and outdoor scenarios.

\section{Methods}
The pipeline of our system is shown in Fig.~\ref{fig:system}. We propose MCN-SLAM, a collaborate multi-agent SLAM system. The input of this multi-agent system is RGB-D frames $\{I_i, D_i\}_{i=1}^M$ with known camera intrinsic $K \in R_{
3\times3}$.  Our model predicts multi-agents camera poses $\{R_i|t_i\}^M_{i=1}$, color $\mathbf{c}$, and implicit truncated signed distance function (TSDF) representation.
The system
consists of four main modules: (i) local (single-robot) mapping (Sec.~\ref{Sec: mapping}),
(ii) distributed camera tracking via PGO (Sec.~\ref{Sec: tracking}), (iii) intra-to-inter loop closure (Sec.~\ref{Sec:loop}) (iv) sub-map fusion (Sec.~\ref{Sec:submap}).  Among these modules, inter-loop closure and submap fusion are the ones that involve communication between different robots.
We elaborate on the entire pipeline of our system in the following subsections.

\subsection{Hybrid Scene Representation}
\label{Sec: mapping} 
Voxel grid-based architectures~\cite{niceslam} are the mainstream in NeRF-based SLAM system. However, they face challenges with cubic memory growth and real-time performance in large-scale environments. To enhance the scene representation capability, we design a novel hybrid scene representation method with planar-grid-coordinate encoding network. The planar encoding network employs tri-plane feature encoding for fast convergence, the grid encoding network employs hashgrid encoding network for high-fidelity reconstruction, while the coordinate encoding uses one-blob encoding with the MLP for smoothness and hole-filling capabilities. Specifically, we design a dual-scale planar-grid encoding network, consisting of both coarse and fine levels to model the scene. We use low-resolution tri-planes network to capture the low-frequency components of the scene. The network can be formulated as:
\begin{align}
\boldsymbol{T}_g(x) =T_{g-x y}^c(x)+T_{g-x z}^c(x)+T_{g-y z}^c(x)
\end{align}
where $\boldsymbol{T}_g(x)$ denotes the tri-plane geometry features. $x$ is the world coordinate of sampling points. $\{T^c_{g-x y}(\cdot),T^c_{g-x z}(\cdot),T^c_{g-y z}(\cdot)\}$ represent the three coarse feature planes.  Similarly, the color tri-plane $\boldsymbol{T}_c(x)$ is defined as:
\begin{align}
    \boldsymbol{T}_c(x) & =T_{c-x y}^c(x)+T_{c-x z}^c(x)+T_{c-y z}^c(x)
\end{align}
where $T^c_c(x),T^f_c(x)$ denote the coarse feature from color tri-planes. $\{T^c_{c-x y}(\cdot),T^c_{c-x z}(\cdot),T^c_{c-y z}(\cdot)\}$ represent the three coarse color feature planes.

Then we use high-resolution voxel grid to model the high-frequency components of the scene. $\boldsymbol{V}_g(x)$, $\boldsymbol{V}_c(x)$ represents the geometry and color grid feature. 
For a sample point $x$, we use bi-linear interpolating the nearest neighbors on  feature plane and tri-linearly interpolating on hashgrid. Then, we concatenate the interpolated tri-plane feature and hashgrid feature as the scene encoding feature. Unlike \cite{niceslam,eslam}, which decode the features directly, our approach leverages a coordinate-based scene representation to model the inherent coherence priors required for effective hole filling.
We use a geometry decoder to predict SDF value $\phi_{ \boldsymbol{g}}(x)$ and a latent feature vector $\mathbf{z}$:
\begin{equation}
f_g\left(\gamma(x),  \boldsymbol{T_g}(x),\boldsymbol{V}_g(x)\right) \mapsto(\mathbf{z}, \phi_{ \boldsymbol{g}}(x))
\end{equation}
where $\gamma(\mathbf{x})$ represents coordinate position encoding. We use one-blob encoding~\cite{coslam} instead of embedding spatial coordinates into multiple frequency bands. Finally, we use a color decoder to predict the RGB value:
\begin{equation}
f_c\left(\gamma(x), \mathbf{z}, \boldsymbol{T_c}(x),\boldsymbol{T_g}(x),\boldsymbol{V}_c(x),\boldsymbol{V}_g(x) \right) \mapsto \phi_{\boldsymbol{a}}(x)
\end{equation}
$\phi_{\boldsymbol{a}}(x)$ represents the color of the sample points. Combining the MLP with the planar-grid scene representation, our carefully designed network architecture achieves accurate and smooth surface reconstruction, efficient memory use, and hole filling performance.

\begin{figure*}[h]
    \centering
     
     \includegraphics[width=\linewidth]{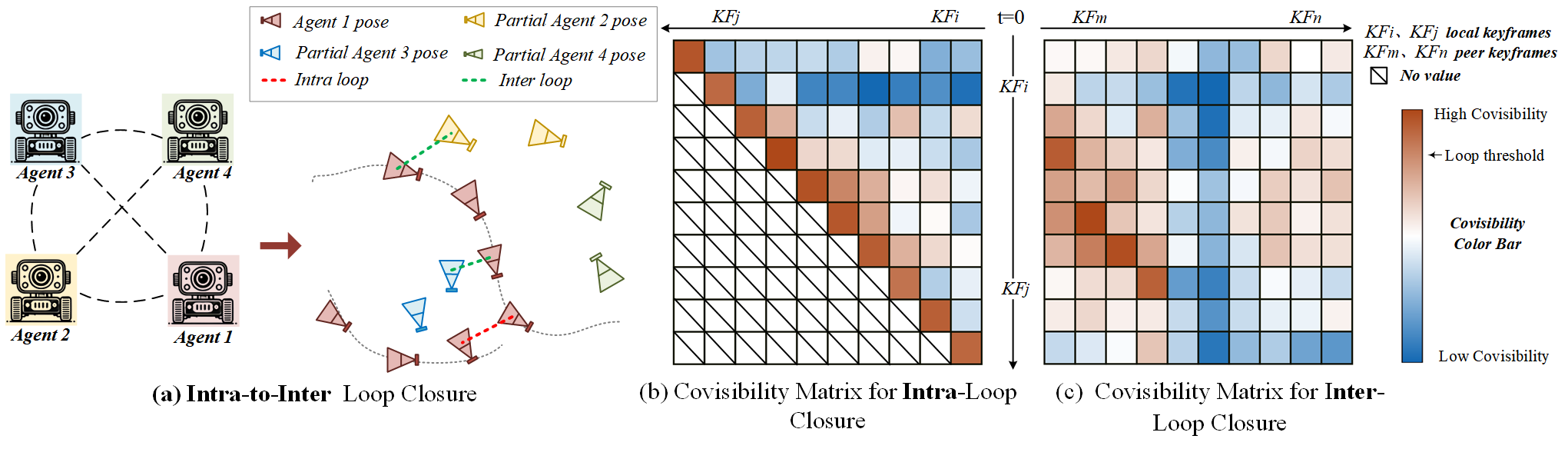}
    \caption{(a) presents the reconstruction of multi-agent pose graph with intra-to-inter loop closure. (b) and (c) present the covisibility matirx of intra loop closure and inter loop closure. In Figure (b), the horizontal and vertical axes represent the keyframes of the local (single-agent) system, while the horizontal and vertical axes in figure (c) represent the keyframes of the local and peer agents. }
    \label{fig:loop}
     
\end{figure*}

\noindent \textbf{Normalized Triplane Scene Representation} Different from the reconstruction of small indoor objects, large-scale and unbounded scenes impose higher demands on feature representation and resolution. The original feature grid method is unscalable for large-scale scenarios due to its cubically growing complexity. Furthermore, for scenes spanning hundreds of meters, the resolution of triplane features remains inadequate. To address this, we propose a normalized triplane representation. Specifically, we normalize the triplane network of each local scene representation to the range [-1,1]. For each submap in the local map sequence, we compute the norm of bound and define a bounded scale value, ensuring a stable and scalable representation for large-scale environments.

\noindent \textbf{Differentiable Rendering}  
We use a differentiable rendering process to integrate the predicted density and colors from our local scene representation. We determine a ray $r(t)=\mathbf{o}+t\mathbf{d}$ whose origin is at the camera center of projection $o$, ray direction $r$. We uniformly sample $K$ points. The sample bound is within the near and far planes $t_k\in [t_n,t_f]$, $k \in \{1,\dots,K\}$ with depth values $\{\mathbf{d_1}, \dots, \mathbf{d_K}\}$ and predicted colors $\{\mathbf{c_1}, . . . , \mathbf{c_K}\}$. For rays with valid depth measurements \( d \), we additionally sample $K_{surface}$ near-surface points uniformly in the interval \( [d - \delta_d,\, d + \delta_d] \), where \( \delta_d \) is a small offset around the measured depth. For all sample points along rays, we query TSDF $\phi_{ \boldsymbol{g}}(x_k)$ and raw color $\phi_{\boldsymbol{a}}(x_k)$ from our
networks and use the SDF-Based rendering approach to convert SDF values to volume densities:
\begin{equation}
\boldsymbol{\sigma}\left(x_k\right)=\frac{1}{\beta} \cdot \operatorname{Sigmoid}\left(\frac{-\phi_{\boldsymbol{g}}\left(x_k\right)}{\beta}\right)
\end{equation}
where $\beta \in \mathbb{R} $ is a learnable parameter that controls the sharpness of the surface boundary.
Then we define the termination probability $w_k$, depth $\hat{\boldsymbol{d}}$, and color $\hat{\boldsymbol{c}}$ as:
\begin{equation}
\begin{gathered}
w_k=\exp \left(-\sum_{m=1}^{n-1} \boldsymbol{\sigma}\left(x_m\right)\right)\left(1-\exp \left(-\boldsymbol{\sigma}\left(x_k\right)\right)\right) \\
\hat{\boldsymbol{c}}=\sum_{k=1}^N w_k \boldsymbol{\phi}_{\boldsymbol{a}}\left(x_k\right) \quad \text { and } \quad \hat{\boldsymbol{d}}=\sum_{k=1}^N w_k t_k
\end{gathered}
\end{equation}
We employ differentiable rendering to model the  color, depth, and SDF of the scene, enabling high-accuracy online reconstruction.

\subsection{Distributed Camera Tracking}
\label{Sec: tracking}
In multi-agent collaboration SLAM system, distributed pose estimation are crucial under communication bandwidth restriction and accumulation of pose errors. In this work, we propose a distributed camera tracking framework involves two stages. In the first stage, we employ RAFT-based~\cite{raft} visual features to estimate the initial trajectory within the local frame of each robot. Additionally, we design an intra-loop closure method based on RAFT features to mitigate pose drifts and enhance global consistency.
In the second stage, we introduce an inter-loop closure strategy with a consistency loss. We first align the pairwise coordinate frames from different robots $\alpha$, $\beta$, transforming each robot’s local frame into a shared global frame. Subsequently, based on each robot's local odometry, all frames are aligned to their respective local reference frames with rendered RGB, depth, and SDF loss. During the optimization process in this stage, we fix the poses of the local reference frames and optimize only the remaining frames. This procedure can be performed in a distributed fashion, with computation occurring independently within each robot. 

In local frame pose estimation of each robot, we take a live RGB-D video stream as input. Our system first applies a recurrent update operator based on RAFT~\cite{raft} to compute the optical flow of each new frame compared to the last keyframe. Each input image is processed by a feature extraction network. The network consists of 6 residual blocks, and 3 downsampling layers, which can produce dense feature map. We use two networks: feature network to build the set of correlation volumes, and context network for the update operator. 

We define a lookup operator which indexes the correlation using a grid with radius r, $L_r: \mathbb{R}^{H \times W \times H \times W} \times \mathbb{R}^{H \times W \times 2} \mapsto \mathbb{R}^{H \times W \times(r+1)^2}$. The lookup operator takes an $H \times W$ grid of coordinates as input and values from correlation volume using bilinear interpolation. The lookup operator is applied to each correlation volume in the pyramid and the final feature vector is computed by concatenating the results as each level. The update operator for optical flow is a $3\times 3$ convolutional GRU with hidden state $\mathbf{h}$. The operator updates the current depth and pose estimates through retraction on the SE3 manifold and vector addition:
\begin{align}
\{\mathbf{R}|\mathbf{t}\}^{(k+1)} &=\operatorname{Exp}\left(\Delta \boldsymbol{\xi}^{(k)}\right) \circ \{\mathbf{R}|\mathbf{t}\}^{(k)}, \\
\mathbf{d}^{(k+1)}&=\Delta \mathbf{d}^{(k)}+\mathbf{d}^{(k)}
\end{align}
We use the current estimated of poses and depths to estimate the flow:
\begin{equation}
    \mathbf{p}_{i j}= \Pi_c\left(\{\mathbf{R}|\mathbf{t}\}_{i j} \circ \Pi_c^{-1}\left(\mathbf{p}_i, \mathbf{d}_i\right)\right)
\end{equation}
\begin{equation}
    \{\mathbf{R}|\mathbf{t}\}_{i j}=\{\mathbf{R}|\mathbf{t}\}_j \circ \{\mathbf{R}|\mathbf{t}\}_i^{-1}
\end{equation}
Here $\Pi_c$ is the camera model mapping a set of 3D points onto the image, and $\Pi_c^{-1}$ is the inverse projection function.
If the average flow is larger than a pre-defined $ \tau_{k}$, a new keyframe is created and added to the global keyframe database for further refinement. We use the set of keyframes $\{KF_i\}_{i=1}^N$ to create a keyframe-graph structure $(\mathcal{V}, \mathcal{E})$, which represents the co-visibility between frames. An edge $(i,j)\in \mathcal{E}$ means $I_i,I_j$ have overlapping fields of view which shared points. After each pose update, we can recompute the visibility to update the frame graph. The keyframe graph is built dynamically with the system operation. 

Afterward, we use the differentiable Dense Bundle Adjustment (DBA) layer to solve a non-linear squares optimization problem to correct the camera pose $\{R_i|t_i\}^M_{i=1}$. 
\begin{equation}
\begin{split}
\mathbf{E}\left(\{\mathbf{R}|\mathbf{t}\}, \mathbf{d}\right)=\sum_{(i, j) \in \mathcal{E}}\left\|\mathbf{p}_{i j}^*- \Pi_c(\mathbf{\{R|t\}}_{i j} \circ  \Pi_c^{-1} \right . \\ \left .
\left(\mathbf{p}_i,    \mathbf{d}_i \right) ) \right\| _{\Sigma_{i j}}^2  + \alpha \sum \left\|\mathbf{\hat{d_i}}-\mathbf{D_i}\right\|^2 
\end{split}
\label{eq:dba}
\end{equation}
where $\Pi_c,\Pi_c^{-1}$ are the projection and back-projection functions, $\|\cdot\|_{\Sigma}$ and $\Sigma_{i j}=\operatorname{diag} \mathbf{w}_{i j}$ denotes the Mahalanobis distance which weights the error terms based on the confidence weights $\mathbf{w}_{ij}$.  $\mathbf{p_i}$ is the 2D pixel position from keyframe $KF_i$. $\{\mathbf{R}|\mathbf{t}\}_{ij}$ is the pose transformation from $KF_i$ to $KF_j$. We use $\mathbf{p}_{i j}^*$ to present the estimated flow. $\alpha$ denotes the weight of depth residual. This loss function states that we want to optimize the camera pose and per-pixel depth to maximize the compatibility with flow $\mathbf{p}_{i j}^*$ predicted by the recurrent update operator. We use local parameterization to linearize Eq.~\ref{eq:dba} and use the Gauss-Newton algorithm solve for updates $(\Delta \boldsymbol{\xi}, \Delta \mathrm{d})$. This parametrization of the structure leads to an extremely efficient way of solving the dense
BA problem, which can be formulated as a linear least squares problem by linearizing the system of equations into the familiar cameras/depths arrow-like block-sparse Hessian $H \in \mathbb{R}^{(c+p) \times(c+p)}$, where c and p are the dimensionality of the cameras and the points. Then the system can be solved efficiently using the Schur complement with the pixelwise damping factor $\lambda$ added to the depth block:

\begin{equation}
    \left[\begin{array}{cc}
\mathbf{B} & \mathbf{E} \\
\mathbf{E}^T & \mathbf{C}
\end{array}\right]\left[\begin{array}{c}
\Delta \boldsymbol{\xi} \\
\Delta \mathbf{d}
\end{array}\right]=\left[\begin{array}{c}
\mathbf{v} \\
\mathbf{w}
\end{array}\right]
\end{equation}
\begin{equation}
    \begin{aligned}
& \Delta \boldsymbol{\xi}=\left[\mathbf{B}-\mathbf{E C}^{-1} \mathbf{E}^T\right]^{-1}\left(\mathbf{v}-\mathbf{E C}^{-1} \mathbf{w}\right) \\
& \Delta \mathbf{d}=\mathbf{C}^{-1}\left(\mathbf{w}-\mathbf{E}^T \Delta \boldsymbol{\xi}\right)
\end{aligned}
\end{equation}
where $\mathbf{C}$ is diagonal and can be cheaply $\mathbf{C}^{-1}=1 / \mathbf{C}$. The DBA layer is integrated into the computation graph, allowing backpropagation to occur through the layer during training. 
 \begin{figure*}[h]
    \centering
     
     \includegraphics[width=\linewidth]{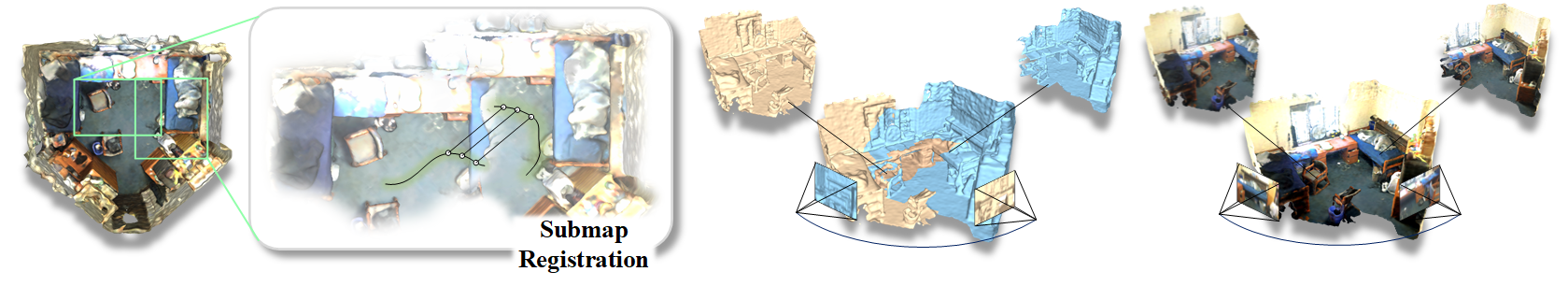}
     
    \caption{
    The multi-implicit-submap fusion for multi-agent SLAM system. Two submaps $M_1$, $M_2$ with their subvolumes and
keyframes are shown. We demonstrate the fusion process of two submaps, where pose and submap bundle adjustment (BA) optimization is performed through loop detection and inter loop closure between agents, ultimately registering the submaps generated by different agents. }
    \label{fig:submap}
     
\end{figure*}

\subsection{Intra-to-Inter Loop Closure}

Most neural implicit SLAM frameworks suffer from accumulation of pose drifts and distortion in the reconstruction. Their tracking networks are performed via minimizing rgb loss functions with
respect to learnable parameters $\theta$ to estimate the relative pose matrix $\{R_i |t_i \} \in \mathbb{S E}(3)$.
With the growing cumulative error $ \varepsilon $ of pose estimation, those methods result in failure in large-scale indoor scenes and long videos. Concurrently, during multi-agent collaborative mapping in large-scale environments, the accumulation pose error of multiple agents results in a rapid degradation of the submaps. To address these problems, we propose a novel intra-to-inter loop closure method, which can eliminate pose drift and achieve global consistency across multiple submaps. We illustrate the schematic diagram of the intra-to-inter loop closure in Fig.~\ref{fig:loop}.

\begin{figure*}[!t]
    \centering

      \includegraphics[width=\linewidth]{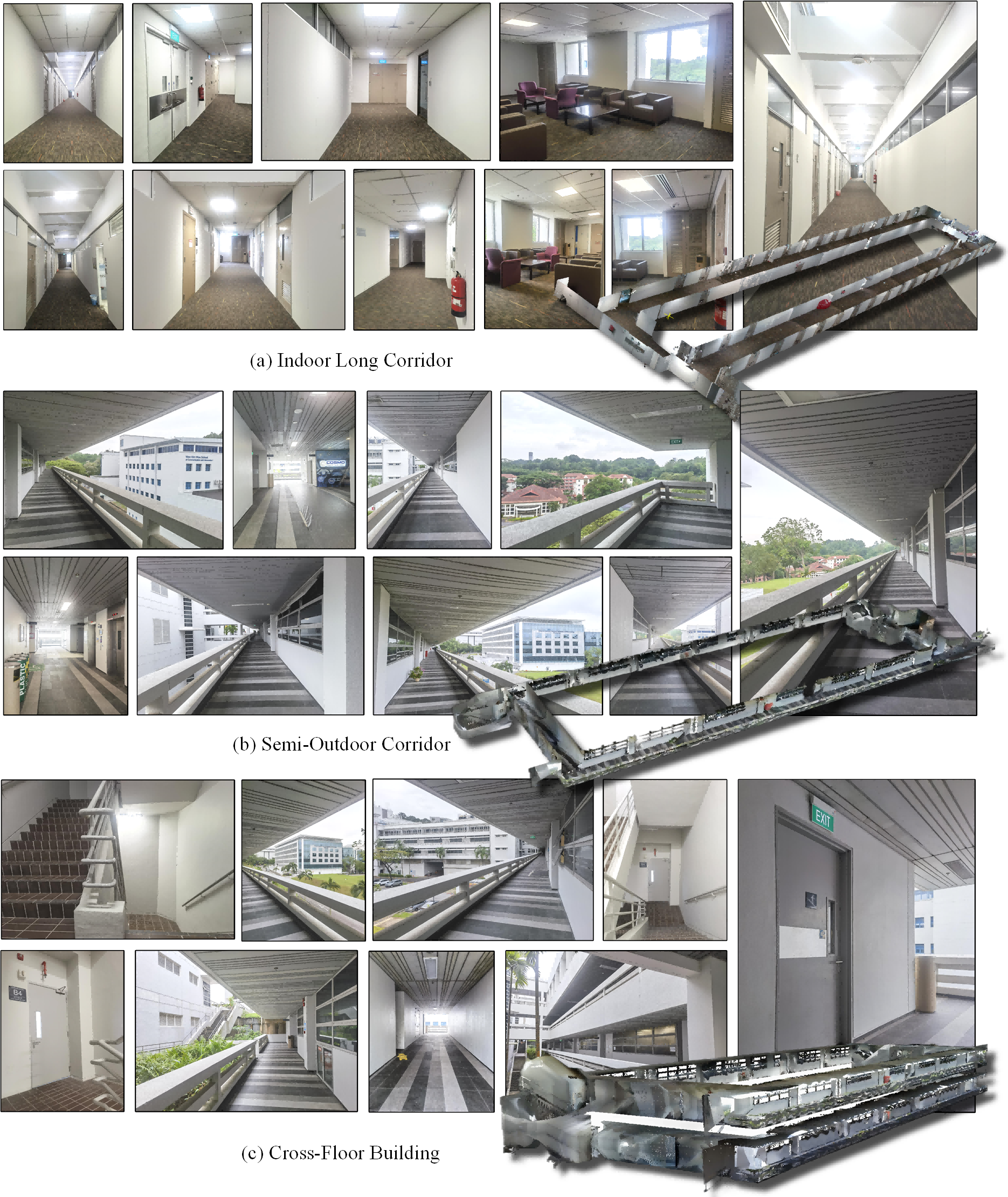}
     
    \caption{
    Visualization of examples from our dense SLAM dataset (DES) in different scenes: (a) Indoor Long corridor, (b) Semi-Outdoor Corridor, (c) Cross-Floor Building.
}
    \label{fig:dataset1}
     
\end{figure*}

\begin{figure*}[!t]
    \centering
     
     \includegraphics[width=\linewidth]{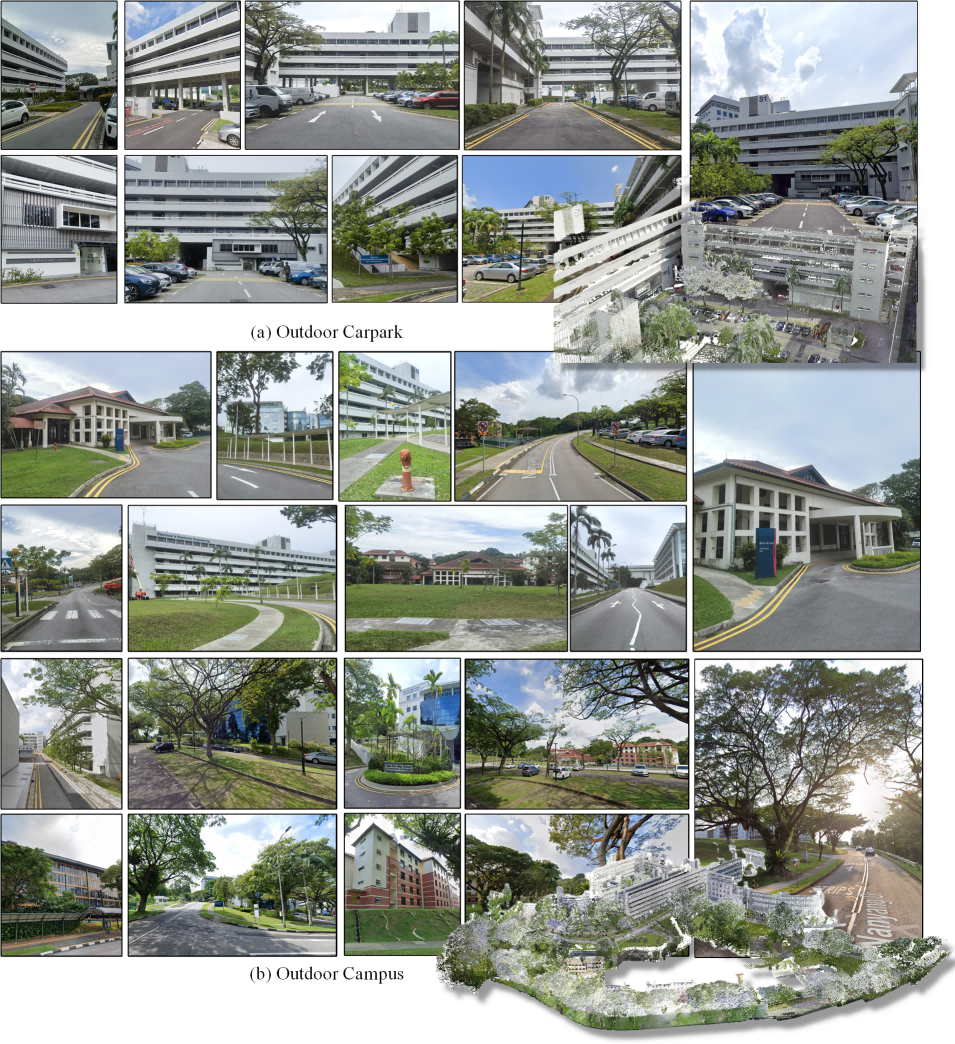}
     
    \caption{
    Visualization of examples from our dense SLAM dataset (DES) in different scenes: (a) Outdoor Carpark, (b) Outdoor Campus.
}
    \label{fig:dataset2}
     
\end{figure*}

\noindent \textbf{Intra-Loop Closure} For single agent (intra) loop closure, we build the local keyframe-graph $(\mathcal{V}, \mathcal{E}) $ with two steps: (i) detect and select keyframe pairs with high covisibility $\tau_{cov}$ in the local most recent keyframes $N_{local}$ (ii) detect loop closure between local keyframes and historical keyframes outside the local window. Accordingly, we compute a covisibility matrix of local keyframes of size $N_{local} \times N_{KF}$ for local loop closure. We also compute a covisibility matrix of local historical keyframes  of size $N_{KF} \times N_{KF}$ for global loop closure in each robot, as shown in Fig.~\ref{fig:loop} (b). The covisibility is represented by the mean rigid flow between keyframe pairs using efficient back-projection. Those keyframe pairs with low covisibility (mean flow higher than thershold ($\tau_{cov}$) are filtered out. We build edges for these keyframe pairs. We suppress the possible neighboring edges between $\{KF_k\}_{k=i-r_{local}}^{i+r_{local}} \to \{KF_k\}_{k=i-r_{local}}^{j+r_{local}}$, where $r_{local}$ denotes a temporal radius. For local loop closure, the number of edges in the graph is linear to $N_{local}$ with an upper bound $N_{local}\times N_{local}$. The number of edges of the global loop closure is linear to $N_{KF}$ with an upper bound $N_{KF}\times N_{KF}$. We also suppress the redundant neighboring edges with radius $r_{global}$. By constructing a pose graph using historical frames, we can leverage all information of each agent to optimize the pose of the current frame. Through neighborhood suppression and covisibility filtering, we limit the number of edges in the keyframe-graph and ensure the efficiency of optimization of intra-loop closure.
Intra-loop closure effectively integrates local information which greatly improves the accuracy of initial pose estimation of each agent in large-scale indoor scenes.

\noindent \textbf{Inter-Loop Closure} Following the first stage, robots use visual feature to obtain optimal trajectory estimates. Then, in the second stage, we use multi-agent inter-loop closure to register pose and submaps from different robots using all odometric measurements and putative loop closures. 
 We formulate the problem to optimize the pose variables from different robots:
 \begin{equation}
\begin{aligned}
 \min _{\substack{\{R|t\}_{\alpha_i} \in \mathrm{SE}(3) \\
\forall \alpha \in \mathcal{R}, \forall i}} \underbrace{\sum_{\alpha \in \mathcal{R}} \sum_{i=1}^{n_\alpha-1} \mathcal{L}\left(\{R|t\}_{\alpha_i}, \{R|t\}_{\alpha_{i+1}}\right)}_{\text {odometry }}+ \\
 \underbrace{\sum_{\left(\alpha_i, \beta_j\right) \in L} \mathcal{L}\left(r_{\beta_i}^{\alpha_i}\left(\{R|t\}_{\alpha_i}, \{R|t\}_{\beta_j}\right)\right),}_{\text {loop closures }}
\end{aligned}
\end{equation}
where $\{R|t\}_{\alpha_i}$ denotes the ith pose of robot $\alpha$ in the global frame. $\mathcal{R}=\{\alpha,\beta,\dots\}$ denotes the set of robots, $n_{\alpha}$ is the total number of poses of robot $\alpha$.

We use keyframe visual descriptors to detect the loop  of different agents. Once the covisibility value of the keyframe pairs is lower than the thershold $\tau_{cov}$, we perform inter loop closure on these frames. We present the covisibility matrix in Fig.~\ref{fig:loop}(c).  For the robot $\alpha$, $\beta$ and the detected frames $A$, $B$, we set the odometric estimates of pose i and pose j as $\{\mathbf{R} |\mathbf{t} \}_{\alpha_i}^A, \{\mathbf{R} |\mathbf{t} \}_{\beta_j}^B  \in \mathbb{S E}(3)$. Then, the relative transformation between keyframe pairs $A$, $B$ can be formulate as:
\begin{equation}
    \{\mathbf{R} |\mathbf{t} \}_{B_{ij}}^A \triangleq \{\mathbf{R} |\mathbf{t} \}_{\alpha_i}^A \{\mathbf{R} |\mathbf{t} \}_{\beta_j}^{\alpha_i}\left( \{\mathbf{R} |\mathbf{t} \}_{\beta_j}^B \right)^{-1}
\end{equation}
where the subscript of $\{\mathbf{R}_i |\mathbf{t}_i \}_{B_{ij}}^A$ indicates that this estimate is computed using inter loop closure (i,j). In order to obtain a reliable estimate of the true relative transformation, we formulate the problem:
\begin{equation}
 \{\mathbf{R} |\mathbf{t} \}_{B_{ij}}^A \in \underset{ \{\mathbf{R} |\mathbf{t} \} \in \operatorname{SE}(3)}{\arg \min } \sum_{(i, j) \in L_{\alpha, \beta}} \mathcal{L}\left(r_{i j}(\{\mathbf{R} |\mathbf{t} \})\right)
\end{equation}
where $\mathcal{L}$ denotes the loss function. $L_{\alpha, \beta}$ is the set of inter-robot loop closure between robot $\alpha$ and $\beta$. We design a consistency loss to register the pose and submaps from different robots:
\begin{align}
    \mathcal{L}_{lc}(\{\mathbf{R} |\mathbf{t} \})&= 
\frac{1}{n} \sum_{\left\{A, B\right\}}^{n} \left( \hat{\boldsymbol{c}_A}-\hat{\boldsymbol{c}_B}\right)^2 \\
\mathcal{L}_{ld}(\{\mathbf{R} |\mathbf{t} \})&= 
\frac{1}{n} \sum_{\left\{A, B\right\}}^{n} \left( \hat{\boldsymbol{d}_A}-\hat{\boldsymbol{d}_B}\right)^2
\end{align}
where $\hat{\boldsymbol{c}_A},\hat{\boldsymbol{c}_B}$ are the rendered color image from agent $\alpha,\beta$, while $\hat{\boldsymbol{d}_A},\hat{\boldsymbol{d}_B}$ denote the rendered depth. 
After that, we fix the poses of the aligned keyframe pairs and construct a rendering loss that incorporates color, depth, and signed distance function (SDF) terms. This loss is used to jointly optimize the poses of the remaining frames within each robot, aligning them consistently to the global world coordinate frame. Our designed inter-loop closure effectively aligns the trajectories and maps across agents. 

\label{Sec:loop}

\subsection{Online Distillation For Submap Fusion}
\label{Sec:submap}
 The sensor observations of multiple agents are local and different, which leads to discrepancies of the final map. In order to achieve global consistency and consensus on scene reconstruction among all agents, two critical components are required: (1) aligning the poses of submaps to ensure global consistency. In Section~\ref{Sec:loop}, we use intra-to-inter loop closure to align the poses of different submaps. (2) fusing the implicit submaps. Unlike traditional point cloud-based maps, implicit map fusion is particularly challenging, especially in distributed systems. 

 Due to communication bandwidth constraints, our peer-to-peer communication only exchanges network parameters and keyframe poses. To this end, we propose a novel multi-agent distillation method to fuse submaps into a global map. We calculate the overlapping region boundaries of different agents using the depth input. Then, we use the poses of the overlapping region keyframe $\{KF\}_{i=1}^m$ to render the RGB and depth keyframe images from different agents. This allows the networks of two different agents to merge their parameters solely by comparing the rendered RGB $\mathbf{\hat{c}}$, depth images $\mathbf{\hat{d}}$, and sdf value $\mathbf{\hat{s}}$ with distillation loss $\mathcal{L}_{\alpha-\beta}$:
 \begin{equation}
 \begin{aligned}
     \mathcal{L}_{\alpha-\beta}=\frac{1}{m} \sum_{\alpha,\beta \in \{KF \}}^m\left( \left(\hat{\mathbf{c}}_\alpha-\mathbf{\hat{c}}_\beta \right)^2 + \left(\mathbf{\hat{d_\alpha}}-\mathbf{\hat{d}_\beta}\right)^2+ \right . \\
     \left . \left(\mathbf{\hat{s_\alpha}}-\mathbf{\hat{s}_\beta}\right)^2 \right)
     \end{aligned}
 \end{equation}
 We use this distillation loss to optimize the network parameters of agent $\alpha$ and $\beta$. We learn the scene parameters to ensure that the overall map scene representation parameters remain globally consistent, shown in Fig.~\ref{fig:submap}.
 
\subsection{Distributed Scene Reconstruction Optimization}
Our mapping thread is performed via minimizing our objective functions with respect to network parameters $\theta$ and camera parameters $\{\mathbf{R}|\mathbf{t}\}$.
The color and depth rendering losses are used in our mapping thread:
\begin{equation}
\mathcal{L}_{c}=\frac{1}{N} \sum_{i=1}^N\left(\hat{\mathbf{c}}_i-\mathbf{C_i}\right)^2,\quad
\mathcal{L}_d=\frac{1}{\left|R_i\right|} \sum_{i \in R_i}\left(\mathbf{\hat{d_i}}-\mathbf{D_i}\right)^2
\end{equation}

where $R_i$ is the set of rays that have a valid depth observation. In addition, we design SDF loss, free space loss for our mapping thread. Specifically, for samples within the truncation region, we leverage the depth sensor measurement to approximate the signed distance field:
\begin{equation}
    \mathcal{L}_{s d f}=\frac{1}{\left|R_i\right|} \sum_{r \in R_i} \frac{1}{\left|X_r^{t r}\right|} \sum_{x \in X_r^{t r}}\left(\phi_{\boldsymbol{g}}(x)\cdot T-(\mathbf{D_i}-\mathbf{d})\right)^2
\end{equation}
where $X_r^{t r}$ is a set of points on the ray r that lie in the truncation region, $|\mathbf{D_i} - \mathbf{d}| \le  tr$. We differentiate the weights of points that are closer
to the surface $X_r^{tm}=\{x|x\in|\mathbf{D_i} - \mathbf{d}| \le  0.4tr\}$ from those that are at the tail of the truncation region $X_r^{tt}$  in our SDF loss. 
\begin{equation}
    \mathcal{L}_{sdf_m}=\mathcal{L}_{sdf}\left(X_r^{tm}\right), \quad  \mathcal{L}_{sdf_t}=\mathcal{L}_{sdf}\left(X_r^{tt}\right)
\end{equation}
For sample points that are far from the surface $|D_i - d| \ge  T$: 
\begin{equation}
    \mathcal{L}_{f s}=\frac{1}{|R_i|} \sum_{r \in R_i} \frac{1}{\left|X_r^{f s}\right|} \sum_{x \in X_r^{f s}}\left(\phi_{\boldsymbol{g}}(x)-1\right)^2
\end{equation}
This loss can force the SDF prediction value to be the truncated distance tr.

\begin{table*}[ht]
\centering
\scalebox{0.72}{
\setlength{\tabcolsep}{1.3mm}{
\begin{tabular}{lcccccccccccc}
\toprule
\multirow{2}{*}{Dataset} & \multirow{2}{*}{Year} & \multicolumn{2}{c}{Scenes}               & \multicolumn{4}{c}{Sensor Modalities} & \multicolumn{2}{c}{GroundTruth} & \multicolumn{2}{c}{GT Method} & \multirow{2}{*}{Description}                                                                                                                                                                      \\
                         &                       & Ind.            & Out.              & RGB    & Depth    & Lidar    & IMU   & Pose          & 3D Map          & Pose                & 3D Map  &                                                                                                                                                                                                   \\ \midrule
\rowcolor{C7!50}Tanks and Temples~\cite{tanks}        & ToG'17                & \color{ForestGreen}{\cmark}     & \color{BrickRed}{\xmark}                     & \color{ForestGreen}{\cmark}       & \color{BrickRed}{\xmark}         & \color{BrickRed}{\xmark}         & \color{BrickRed}{\xmark}      & \color{ForestGreen}{\cmark}              & \color{ForestGreen}{\cmark}                &  ICP   & Laser Scanner         & High-quality Object-centric Scenes                                                                                                                                                                \\
\rowcolor{C7!50}RealEstate10k~\cite{RealEstate10k}            & ToG'18                & \color{ForestGreen}{\cmark}     & \color{ForestGreen}{\cmark}                      & \color{ForestGreen}{\cmark}       & \color{BrickRed}{\xmark}         & \color{BrickRed}{\xmark}         & \color{BrickRed}{\xmark}      & \color{ForestGreen}{\cmark}              & \color{BrickRed}{\xmark}              & COLMAP              &  \color{BrickRed}{\xmark}       & Video clips on YouTube captured  from a moving camera                                                                                                   \\
\rowcolor{C7!50}NeRF-LLFF~\cite{llff}                & SIG'19           & \color{ForestGreen}{\cmark}     & \color{BrickRed}{\xmark}                      & \color{ForestGreen}{\cmark}       & \color{BrickRed}{\xmark}         & \color{BrickRed}{\xmark}         & \color{BrickRed}{\xmark}      & \color{ForestGreen}{\cmark}              & \color{BrickRed}{\xmark}              & COLMAP              &   \color{BrickRed}{\xmark}      & Images of various indoor scenes and objects.                                                                                                                                                      \\
\rowcolor{C7!50}ACID~\cite{acid}                     & ICCV'21               & \color{BrickRed}{\xmark}     & \color{ForestGreen}{\cmark}                      & \color{ForestGreen}{\cmark}       & \color{BrickRed}{\xmark}         & \color{BrickRed}{\xmark}         & \color{BrickRed}{\xmark}      & \color{ForestGreen}{\cmark}              & \color{BrickRed}{\xmark}        & COLMAP              &    \color{BrickRed}{\xmark}     & A dataset of aerial landscape videos                                                                                                  \\
\rowcolor{C7!50}Mip-NeRF 360~\cite{mipnerf360}             & CVPR'22               & \color{ForestGreen}{\cmark}      & \color{ForestGreen}{\cmark}                      & \color{ForestGreen}{\cmark}       & \color{BrickRed}{\xmark}         & \color{BrickRed}{\xmark}         & \color{BrickRed}{\xmark}      & \color{ForestGreen}{\cmark}              & \color{BrickRed}{\xmark}                & COLMAP              &  \color{BrickRed}{\xmark}       & Outdoor and indoor scenes with 360-degree scene perspectives                                                                                  \\
\rowcolor{C7!50}LocalRF~\cite{localrf}                  & CVPR'23               & \color{BrickRed}{\xmark}     & \color{ForestGreen}{\cmark}                      & \color{ForestGreen}{\cmark}       & \color{BrickRed}{\xmark}         & \color{BrickRed}{\xmark}         & \color{BrickRed}{\xmark}      & \color{ForestGreen}{\cmark}              & \color{BrickRed}{\xmark}                 & COLMAP              &  \color{BrickRed}{\xmark}       & Large-scale outdoor dataset, Static hikes                                                                                                                                                         \\ \midrule
\rowcolor{C5!10}MITDARPA~\cite{MITDARPA}                    & IJRR'10               & \color{BrickRed}{\xmark}     & \color{ForestGreen}{\cmark}                     & \color{ForestGreen}{\cmark}       & \color{BrickRed}{\xmark}         & \color{ForestGreen}{\cmark}        & \color{BrickRed}{\xmark}      & \color{ForestGreen}{\cmark}              & \color{BrickRed}{\xmark}                 & GPS            &   \color{BrickRed}{\xmark}      & Urban scenes mapping and localization \\
\rowcolor{C5!10}UTIAS~\cite{utias}                    & IJRR'11               & \color{ForestGreen}{\cmark}     & \color{BrickRed}{\xmark}                     & \color{ForestGreen}{\cmark}       & \color{BrickRed}{\xmark}         & \color{BrickRed}{\xmark}         & \color{BrickRed}{\xmark}      & \color{ForestGreen}{\cmark}              & \color{BrickRed}{\xmark}                 & Motion Capture            &   \color{BrickRed}{\xmark}      & Multi-agent indoor scenes mapping and localization \\
\rowcolor{C5!10}Ford Campus~\cite{ford}                    & IJRR'11               & \color{BrickRed}{\xmark}     &\color{ForestGreen}{\cmark}                     & \color{ForestGreen}{\cmark}       & \color{BrickRed}{\xmark}         & \color{ForestGreen}{\cmark}         & \color{ForestGreen}{\cmark}      & \color{ForestGreen}{\cmark}              & \color{BrickRed}{\xmark}                 & GPS            &   \color{BrickRed}{\xmark}      & Campus scenes mapping and localization \\\rowcolor{C5!10}TUM RGB-D~\cite{tum}                & IROS'12               & \color{ForestGreen}{\cmark}     & \color{BrickRed}{\xmark}                      & \color{ForestGreen}{\cmark}       & \color{ForestGreen}{\cmark}         & \color{BrickRed}{\xmark}         & \color{BrickRed}{\xmark}      & \color{ForestGreen}{\cmark}              & \color{BrickRed}{\xmark}                  & Motion Capture      &  \color{BrickRed}{\xmark}       & Various indoor scenes for mapping and localization \\
\rowcolor{C5!10}KITTI~\cite{kitti}                    & IJRR'13               & \color{BrickRed}{\xmark}     & \color{ForestGreen}{\cmark}                      & \color{ForestGreen}{\cmark}       & \color{BrickRed}{\xmark}         & \color{ForestGreen}{\cmark}         & \color{ForestGreen}{\cmark}      & \color{ForestGreen}{\cmark}              & \color{BrickRed}{\xmark}                 & GNSS/INS            &   \color{BrickRed}{\xmark}      & A city-scale dataset created for autonomous driving research                                                                            \\
\rowcolor{C5!10}NCLT~\cite{nclt}                    & IJRR'16               & \color{BrickRed}{\xmark}     & \color{ForestGreen}{\cmark}                      & \color{ForestGreen}{\cmark}       & \color{BrickRed}{\xmark}         & \color{ForestGreen}{\cmark}         & \color{ForestGreen}{\cmark}      & \color{ForestGreen}{\cmark}              & \color{BrickRed}{\xmark}                 & GNSS/INS            &   \color{BrickRed}{\xmark}      &  Long-term campus scenes mapping and localization \\
                                            
\rowcolor{C5!10}ShapeNet~\cite{shapenet}                 & arXiv'15              & \color{ForestGreen}{\cmark}     & \color{BrickRed}{\xmark}                      & \color{ForestGreen}{\cmark}       & \color{BrickRed}{\xmark}         & \color{BrickRed}{\xmark}         & \color{BrickRed}{\xmark}      & \color{ForestGreen}{\cmark}              & \color{ForestGreen}{\cmark}                  &   Simulation             &   CAD      & A richly-annotated, large-scale dataset of 3D shapes                                                                                                   \\
\rowcolor{C5!10}Euroc~\cite{euroc}                    & IJRR'16               & \color{ForestGreen}{\cmark}     & \color{BrickRed}{\xmark}                      & \color{ForestGreen}{\cmark}       & \color{BrickRed}{\xmark}           & \color{BrickRed}{\xmark}         & \color{ForestGreen}{\cmark}      & \color{ForestGreen}{\cmark}              & \color{BrickRed}{\xmark}                 & Motion Capture            &   \color{BrickRed}{\xmark}      & Various indoor scenes for mapping and localization\\
\rowcolor{C5!10}Oxford RobotCar~\cite{oxfordrobo}                    & IJRR'17               & \color{BrickRed}{\xmark}     & \color{ForestGreen}{\cmark}                     & \color{ForestGreen}{\cmark}       & \color{BrickRed}{\xmark}          & \color{ForestGreen}{\cmark}         & \color{ForestGreen}{\cmark}      & \color{ForestGreen}{\cmark}              & \color{BrickRed}{\xmark}                 & GNSS            &   \color{BrickRed}{\xmark}      & A city-scale dataset created for autonomous driving research \\
\rowcolor{C5!10}ScanNet~\cite{scannet}                  & CVPR'17               & \color{BrickRed}{\xmark}     & \color{BrickRed}{\xmark}                      & \color{ForestGreen}{\cmark}       & \color{ForestGreen}{\cmark}         & \color{BrickRed}{\xmark}         & \color{BrickRed}{\xmark}      & \color{ForestGreen}{\cmark}              & \color{BrickRed}{\xmark}               & SLAM                &  \color{BrickRed}{\xmark}       & A richly-annotated multiple room scenes with semantic label                                                                                            \\
\rowcolor{C5!10}Replica~\cite{replica}                  & arXiv'19              & \color{ForestGreen}{\cmark}     & \color{BrickRed}{\xmark}                      & \color{ForestGreen}{\cmark}       & \color{ForestGreen}{\cmark}         & \color{BrickRed}{\xmark}         & \color{BrickRed}{\xmark}      & \color{ForestGreen}{\cmark}              & \color{BrickRed}{\xmark}             & Simulation          &    Simulation      & Highly photo-realistic 3D indoor scene dataset at room scale                                                                                          \\
\rowcolor{C5!10}Nuscenes~\cite{nuscenes}                 & CVPR'20               &\color{BrickRed}{\xmark}     & \color{ForestGreen}{\cmark}                      & \color{ForestGreen}{\cmark}       & \color{BrickRed}{\xmark}        & \color{BrickRed}{\xmark}         & \color{BrickRed}{\xmark}      & \color{ForestGreen}{\cmark}              & \color{BrickRed}{\xmark}             & GNSS/INS            &  \color{BrickRed}{\xmark}       & A city-scale dataset created for autonomous driving research                                                                                           \\
\rowcolor{C5!10}Waymo~\cite{waymo}                    & CVPR'20               & \color{BrickRed}{\xmark}     & \color{ForestGreen}{\cmark}                      & \color{ForestGreen}{\cmark}       & \color{BrickRed}{\xmark}        & \color{BrickRed}{\xmark}         & \color{BrickRed}{\xmark}      & \color{ForestGreen}{\cmark}              & \color{BrickRed}{\xmark}               & GNSS/INS            &  \color{BrickRed}{\xmark}       & A city-scale dataset created for autonomous driving research                                                                                          \\
\rowcolor{C5!10}UMA-VI~\cite{umavi}                    & IJRR'20               & \color{ForestGreen}{\cmark}     & \color{BrickRed}{\xmark}                      & \color{ForestGreen}{\cmark}       & \color{BrickRed}{\xmark}           & \color{BrickRed}{\xmark}         & \color{ForestGreen}{\cmark}      & \color{ForestGreen}{\cmark}              & \color{BrickRed}{\xmark}                 & Motion Capture            &   \color{BrickRed}{\xmark}      & Various indoor scenes for mapping and localization with Hand-held equipment \\
\rowcolor{C5!10}TartanAir~\cite{tartanair}                & IROS'20               & \color{ForestGreen}{\cmark}     & \color{ForestGreen}{\cmark}                      & \color{ForestGreen}{\cmark}       & \color{ForestGreen}{\cmark}        & \color{ForestGreen}{\cmark}         & \color{ForestGreen}{\cmark}      & \color{ForestGreen}{\cmark}              & \color{BrickRed}{\xmark}                & Simulation          &     \color{BrickRed}{\xmark}     &  A Large-scale dataset with various light conditions,  weather, and moving objects                                                                     \\
\rowcolor{C5!10}NTU VIRAL~\cite{ntu}                & IJRR'22               & \color{BrickRed}{\xmark}     & \color{ForestGreen}{\cmark}                      & \color{ForestGreen}{\cmark}       & \color{BrickRed}{\xmark}        & \color{BrickRed}{\xmark}         & \color{BrickRed}{\xmark}      & \color{ForestGreen}{\cmark}              & \color{BrickRed}{\xmark}               & TLS                 &   \color{BrickRed}{\xmark}       & A dataset of aerial mapping and localization                                                                                                                                                      \\
\rowcolor{C5!10}ScanNet++~\cite{scannet++}                & ICCV'23               & \color{ForestGreen}{\cmark}     & \color{BrickRed}{\xmark}                      & \color{ForestGreen}{\cmark}       & \color{ForestGreen}{\cmark}         & \color{BrickRed}{\xmark}         & \color{BrickRed}{\xmark}      & \color{ForestGreen}{\cmark}              & \color{ForestGreen}{\cmark}               & SLAM                & Laser Scanner        & A large-scale dataset with high-quality and geometry and color of indoor scenes.                                                                    \\ \rowcolor{C5!10}SubT-MRS~\cite{subT}                    & CVPR'24               & \color{BrickRed}{\xmark}     & \color{ForestGreen}{\cmark}                      & \color{ForestGreen}{\cmark}       & \color{BrickRed}{\xmark}         & \color{ForestGreen}{\cmark}         & \color{ForestGreen}{\cmark}      & \color{ForestGreen}{\cmark}              & \color{BrickRed}{\xmark}                 & GNSS/INS            &   \color{BrickRed}{\xmark}      & Mapping and localization under diverse all-weather conditions  \\
\rowcolor{C5!10} MCD~\cite{mcd}                     & CVPR'24              & \color{BrickRed}{\xmark}      & \color{ForestGreen}{\cmark}                      & \color{ForestGreen}{\cmark}       & \color{ForestGreen}{\cmark}        & \color{ForestGreen}{\cmark}        & \color{ForestGreen}{\cmark}      & \color{ForestGreen}{\cmark}              & \color{ForestGreen}{\cmark}            &  MCTR   &  Laser Scanner       & A multi-campus dataset focus on semantic and SLAM
\\\rowcolor{C5!10}MARS-LVIG~\cite{mars}                    & IJRR'24               & \color{BrickRed}{\xmark}     & \color{ForestGreen}{\cmark}                      & \color{ForestGreen}{\cmark}       & \color{BrickRed}{\xmark}         & \color{ForestGreen}{\cmark}         & \color{ForestGreen}{\cmark}      & \color{ForestGreen}{\cmark}              & \color{BrickRed}{\xmark}                 & GNSS/INS            &   \color{BrickRed}{\xmark}      & Aerial downward-looking  SLAM dataset \\ 

\rowcolor{C5!10}Ours                     &               & \color{ForestGreen}{\cmark}     & \color{ForestGreen}{\cmark}                      & \color{ForestGreen}{\cmark}       & \color{ForestGreen}{\cmark}        & \color{ForestGreen}{\cmark}        & \color{ForestGreen}{\cmark}      & \color{ForestGreen}{\cmark}              & \color{ForestGreen}{\cmark}            &  MCTR   &  Laser Scanner       & \begin{tabular}[c]{@{}c@{}}A large-scale indoor and outdoor dataset with a   widerange of sensing \\ modalities and high-accuracy groundtruth for both single-agent and multi-agent\end{tabular} \\ \bottomrule
\end{tabular}}}

\caption{List of commonly used datasets for neural mapping and localization. We provide the first dataset that includes both high-precision trajectory ground truth and 3D mesh ground truth. It covers indoor and outdoor scenes, incorporates multiple sensor modalities, and includes both single-agent and multi-agent sequences.}
 
\label{tab:datasetcompare}
\end{table*}

\begin{table*}[]
\centering
\scalebox{0.87}{             
\setlength{\tabcolsep}{1.5mm}{
\begin{tabular}{lclccclccclcc|ccc}
\toprule
\multirow{3}{*}{Methods} & \multicolumn{12}{c|}{Reconstruction}                                                                                                                                                                                                                         & \multicolumn{3}{c}{Localization}                          \\
                         & \multicolumn{4}{c}{Agent 1}                                                        & \multicolumn{4}{c}{Agent 2}                                                        & \multicolumn{4}{c|}{Global}                                                       & Agent 1           & Agent 2           & Global           \\ \cline{2-16} 
                         & \multicolumn{2}{c}{Acc. $\downarrow$} & Comp. $\downarrow$ & Comp.Ratio(\%) $\uparrow$ & \multicolumn{2}{c}{Acc. $\downarrow$} & Comp. $\downarrow$ & Comp.Ratio(\%) $\uparrow$ & \multicolumn{2}{c}{Acc. $\downarrow$} & Comp. $\downarrow$ & Comp.Ratio(\%) $\uparrow$ & RMSE $\downarrow$ & RMSE $\downarrow$ & RMSE $\downarrow$ \\ \midrule
                         \multicolumn{16}{l}{\cellcolor[HTML]{EEEEEE}{\textit{Single-Agent Methods}}} \\
iMAP~\cite{imap}                  & \multicolumn{2}{c}{4.812}             & 4.519              & 76.378                & \multicolumn{2}{c}{3.663}             & 3.582              & 81.352                & \multicolumn{2}{c}{2.881}             & 4.934              & 80.515                & 3.481             & 3.773             & 4.153             \\
NICE-SLAM~\cite{niceslam}            & \multicolumn{2}{c}{3.035}             & 3.907              & 85.894                & \multicolumn{2}{c}{2.163}             & 2.881              & 90.792                & \multicolumn{2}{c}{2.373}             & 2.645              & 91.137                & 1.657             & 1.767             & 2.503             \\
ESLAM~\cite{eslam}      & \multicolumn{2}{c}{1.838}             & \cellcolor{tabsecond}1.979              & 93.317                & \multicolumn{2}{c}{1.857}             & 1.884              & 94.531                & \multicolumn{2}{c}{2.082}             & 1.754              & 96.014                & 0.593             & 0.579             & 0.657             \\
CoSLAM~\cite{coslam}     & \multicolumn{2}{c}{\cellcolor{tabsecond}1.758}             & 2.153              & 92.894                & \multicolumn{2}{c}{1.949}             & 2.033              & 93.413                & \multicolumn{2}{c}{2.104}             & 2.082              & 93.435                & 1.001             & 0.934             & 1.059             \\
GOSLAM~\cite{goslam}    & \multicolumn{2}{c}{2.931}             & 3.759              & 86.484                & \multicolumn{2}{c}{2.245}             & 2.678              & 90.571                & \multicolumn{2}{c}{2.091}             & 2.415              & 93.174                & 0.334             & 0.451             & 0.404             \\
Point-SLAM~\cite{pointslam}           & \multicolumn{2}{c}{2.324}             & 3.455              & 86.811                & \multicolumn{2}{c}{1.925}             & 3.102              & 90.975                      & \multicolumn{2}{c}{1.935}             & 2.779              & 90.479                & 0.534             & 0.461             & 0.559             \\
PLGSLAM~\cite{plgslam}             & \multicolumn{2}{c}{1.793}             & 2.012              & \cellcolor{tabthird}93.415                & \multicolumn{2}{c}{1.983}             & \cellcolor{tabthird}1.773              & \cellcolor{tabthird}94.669                & \multicolumn{2}{c}{1.884}             & \cellcolor{tabthird}1.543              & \cellcolor{tabthird}96.329                & 0.584             & 0.607             & 0.631             \\
Loopy-SLAM~\cite{loopyslam}        & \multicolumn{2}{c}{1.935}             & 3.107              & 89.894                & \multicolumn{2}{c}{\cellcolor{tabfirst}1.683}             & 2.783              & 91.792                & \multicolumn{2}{c}{\cellcolor{tabfirst}1.593}             & 2.645              & 91.137                     & \cellcolor{tabfirst}0.257             & \cellcolor{tabsecond}0.362             & \cellcolor{tabfirst}0.369             \\\multicolumn{16}{l}{\cellcolor[HTML]{EEEEEE}{\textit{Multi-Agent Methods}}} \\
CP-SLAM~\cite{CP-SLAM}   & \multicolumn{2}{c}{2.271}             & 3.457              & 86.429                & \multicolumn{2}{c}{1.834}             & 2.901              & 90.115                & \multicolumn{2}{c}{1.794}             & 2.455              & 90.394                & 0.613             & 0.639             & 0.673             \\
MNE-SLAM~\cite{mneslam}                 & \multicolumn{2}{c}{\cellcolor{tabsecond}1.731}             & \cellcolor{tabsecond}1.932              & \cellcolor{tabsecond}93.692                & \multicolumn{2}{c}{\cellcolor{tabthird}1.797}             & \cellcolor{tabsecond}1.685              & \cellcolor{tabsecond}95.458                & \multicolumn{2}{c}{\cellcolor{tabthird}1.681}             & \cellcolor{tabsecond}1.509              & \cellcolor{tabsecond}96.957                & \cellcolor{tabthird}0.284             & \cellcolor{tabsecond}0.358             & \cellcolor{tabthird}0.398             \\
MCN-SLAM (Ours)                 & \multicolumn{2}{c}{\cellcolor{tabfirst}1.728}             & \cellcolor{tabfirst}1.929              & \cellcolor{tabfirst}93.712                & \multicolumn{2}{c}{\cellcolor{tabsecond}1.793}             & \cellcolor{tabfirst}1.683              & \cellcolor{tabfirst}95.465                & \multicolumn{2}{c}{\cellcolor{tabsecond}1.679}             & \cellcolor{tabfirst}1.519              & \cellcolor{tabfirst}96.969                & \cellcolor{tabsecond}0.280             & \cellcolor{tabfirst}0.353             & \cellcolor{tabsecond}0.392             \\ \bottomrule
\end{tabular}}}

\caption{
Multi-agent SLAM performance on the Replica dataset \cite{replica}. We evaluate the reconstruction and localization performance of Agent1, Agent2, and global(global scene reconstruction and camera tracking). We present the evaluation results of the reconstruction and trajectory outcomes for different agents, as well as the scene reconstruction and trajectory results of the global map. The results are the average of the scenes in the Replica dataset. Our method outperforms existing methods in surface reconstruction and pose estimation. Best results are highlighted as \colorbox{tabfirst}{first}, \colorbox{tabsecond}{second}, and \colorbox{tabthird}{third}.}
\label{tab:replica}

\end{table*}

\begin{table*}[]
\centering
\scalebox{0.90}{
\setlength{\tabcolsep}{1.5mm}{
\begin{tabular}{lccccccccccccccc}
\toprule
\multirow{3}{*}{Methods} & \multicolumn{12}{c}{Reconstruction}                                                                                                                        \\
                         & \multicolumn{4}{c}{Agent 1}                                                        & \multicolumn{4}{c}{Agent 2}                                                        & \multicolumn{4}{c}{Global}                                                                 \\ \cline{2-16} 
                         & Depth L1 (cm)  & PSNR $\uparrow$ & SSIM $\uparrow$ & LPIPS $\downarrow$ & Depth L1 (cm)& PSNR $\uparrow$ & SSIM $\uparrow$ & LPIPS $\downarrow$ & Depth L1 (cm)& PSNR $\uparrow$ & SSIM $\uparrow$ & LPIPS $\downarrow$  \\ \midrule
                         \multicolumn{16}{l}{\cellcolor[HTML]{EEEEEE}{\textit{Single-Agent Methods}}} \\
iMAP~\cite{imap}                & 3.15       & 23.19             & 0.772              & 0.272   & 3.21             & 23.41             &  0.760              & 0.278      & 3.18          & 23.33             & 0.769              & 0.274                  \\
NICE-SLAM~\cite{niceslam}     & 3.01       & 24.28             & 0.820              & 0.250   & 2.96             & 24.57             &  0.799              & 0.257      & 2.98          & 24.43             & 0.809              & 0.254                  \\
ESLAM~\cite{eslam}       & 1.16              &\cellcolor{tabthird} 28.36            & \cellcolor{tabsecond}0.920              & 0.244     & 1.21              & \cellcolor{tabthird}27.83            & \cellcolor{tabsecond}0.927              & \cellcolor{tabthird}0.248   & 1.19              & 28.06            & \cellcolor{tabthird}0.923              & 0.246                   \\
Co-SLAM~\cite{coslam}     & 1.47             & 26.56              & 0.876              & 0.263   &  1.50           & 26.39            & 0.870             & 0.266   & 1.51            &  26.46            &  0.873              &  0.269                   \\
GO-SLAM~\cite{goslam}   & 3.01       & 23.88             & 0.820              & 0.298   & 2.96             & 23.17             &  0.799              & 0.317      & 3.38          & 23.53             & 0.801              & 0.305               \\
PLGSLAM~\cite{plgslam}
    &\cellcolor{tabthird} 1.14              &\cellcolor{tabthird} 28.36            & 0.919              & \cellcolor{tabthird}0.242     & \cellcolor{tabthird}1.20              & \cellcolor{tabthird}27.83            & 0.925              & \cellcolor{tabthird}0.248   & \cellcolor{tabthird}1.17              & \cellcolor{tabthird}28.16            & 0.922              & \cellcolor{tabthird}0.244                        \\
       \multicolumn{16}{l}{\cellcolor[HTML]{EEEEEE}{\textit{Multi-Agent Methods}}} \\
CP-SLAM~\cite{CP-SLAM}     & 1.18             & 28.16              & 0.896              & 0.249   &  1.23           & 27.59            & 0.910             & 0.258   & 1.31            &  27.96            &  0.911              &  0.261                   \\
MNE-SLAM~\cite{mneslam}     &\cellcolor{tabsecond} 0.83              &\cellcolor{tabsecond} 30.79            &\cellcolor{tabsecond} 0.960              &\cellcolor{tabfirst} 0.202     &\cellcolor{tabsecond} 0.90              &\cellcolor{tabsecond} 30.09            &\cellcolor{tabsecond} 0.955              &\cellcolor{tabsecond} 0.204   &\cellcolor{tabsecond} 0.87              &\cellcolor{tabsecond} 30.41            &\cellcolor{tabsecond} 0.953              &\cellcolor{tabsecond} 0.208  \\
Ours     &\cellcolor{tabfirst} 0.81              &\cellcolor{tabfirst} 30.85            &\cellcolor{tabfirst} 0.962              &\cellcolor{tabfirst} 0.200     &\cellcolor{tabfirst} 0.88              &\cellcolor{tabfirst} 30.25            &\cellcolor{tabfirst} 0.953              &\cellcolor{tabfirst} 0.199   &\cellcolor{tabfirst} 0.84              &\cellcolor{tabfirst} 30.58            &\cellcolor{tabfirst} 0.961              &\cellcolor{tabfirst} 0.204                         \\ \bottomrule
\end{tabular}}}
  
\caption{Multi-agent rendering and  reconstruction performance on the Replica dataset \cite{replica}. We evaluate the rendering performance of Agent1, Agent2, and global(global scene reconstruction and camera tracking). The results are the average of the scenes in the Replica dataset. Our method outperforms existing methods in RGB and depth rendering.   }
   
\label{tab:render}
\end{table*}

\begin{table*}[]
\centering
\scalebox{0.90}{
\setlength{\tabcolsep}{1.5mm}{
\begin{tabular}{lccccccccc}
\toprule
\multirow{2}{*}{Methods} & \multicolumn{3}{c}{Apartment-1}                                                       & \multicolumn{3}{c}{Apartment-2}                                                       & \multicolumn{3}{c}{Apartment-0}                                                       \\
                         & Agent 1   & Agent 2  & Global  & Agent 1   & Agent 2  & Global  & Agent 1   & Agent 2  & Global  \\ 
\multicolumn{1}{c}{}     & \multicolumn{3}{c}{RMSE{[}cm{]}/Mean{[}cm{]}/Median{[}cm{]}}                          & \multicolumn{3}{c}{RMSE{[}cm{]}/Mean{[}cm{]}/Median{[}cm{]}}                          & \multicolumn{3}{c}{RMSE{[}cm{]}/Mean{[}cm{]}/Median{[}cm{]}}                          \\ \midrule \multicolumn{10}{l}{\cellcolor[HTML]{EEEEEE}{\textit{Single-Agent Methods}}} \\

ORB-SLAM3~\cite{orbslam3}              & 4.93/4.65/5.01      & 4.93/4.04/3.80     & 4.93/4.35/4.41     & 1.35/1.05/0.65      & 1.36/1.24/1.11     & 1.36/1.15/0.88              & 0.67/0.58/0.47             & 1.46/1.11/0.79             & 1.07/0.85/0.63    \\
NICE-SLAM~\cite{niceslam}  & 55.4/53.4/50.9    & 21.6/20.3/19.4    & 38.5/36.4/35.9     & 5.70/5.53/5.25      & 2.99/2.79/2.51     & 4.35/4.15/4.01      & 2.17/2.01/1.93     & 2.21/2.03/1.94      & 2.18/2.05/1.97     \\
Co-SLAM~\cite{coslam}      & 2.86/2.74/2.58     & 3.51/3.40/3.24     & 3.19/2.95/2.77      & 1.44/1.32/1.27     & 1.64/1.45/1.39     & 1.54/1.44/1.38      & 0.83/0.74/0.67     & 0.78/0.73/0.0.67     & 0.83/0.75/0.68        \\
ESLAM~\cite{eslam}      & 1.38/1.29/1.17     & \cellcolor{tabfirst}0.95/0.89/0.81     & 1.17/1.07/0.98      & 0.84/0.76/0.69     & 0.75/0.69/0.61     & 0.79/0.73/0.67      & 0.58/0.53/0.49     & 0.95/0.88/0.81     & 0.76/0.70/0.65     \\
GO-SLAM~\cite{goslam}                & 1.45/1.38/1.30      & 1.33/1.21/1.16     & 1.27/1.12/1.01     & \cellcolor{tabthird}0.49/0.45/0.43      & 0.78/0.74/0.70     & \cellcolor{tabthird}0.62/0.58/0.53     & 0.47/0.44/0.41      & \cellcolor{tabthird}0.56/0.52/0.49     & \cellcolor{tabthird}0.53/0.50/0.48     \\
Point-SLAM~\cite{pointslam}     & 1.31/1.23/1.11      & 2.09/1.98/1.77     & 1.63/1.51/1.38     & 0.53/0.47/0.42     & 0.80/0.76/0.73     & 0.68/0.62/0.57     & 0.49/0.46/0.42      & 0.62/0.58/0.55     & 0.58/0.55/0.53          \\
PLGSLAM~\cite{plgslam}                  & 1.33/1.26/1.18     & 1.06/0.99/0.91     & \cellcolor{tabsecond}1.13/1.06/0.98      & 0.82/0.77/0.70     & 0.73/0.68/0.65     & 0.79/0.74/0.71      & 0.56/0.53/0.51     & 0.93/0.87/0.83     & 0.74/0.69/0.66          \\ Loopy-SLAM~\cite{loopyslam}               & \cellcolor{tabfirst}1.19/1.07/0.98     & 1.66/1.53/1.44     & 1.43/1.35/1.26      & 0.55/0.51/0.48     & \cellcolor{tabfirst}0.66/0.61/0.53     & 0.60/0.53/0.48      & \cellcolor{tabsecond}0.46/0.42/0.40     & 0.82/0.79/0.76     & 0.64/0.61/0.58           \\ \multicolumn{10}{l}{\cellcolor[HTML]{EEEEEE}{\textit{Multi-Agent Methods}}} \\
CCM-SLAM~\cite{ccmslam}              & 2.12/1.94/1.74      & 9.31/6.36/5.57     & 5.71/4.15/3.66     & 0.51/0.45/0.40      & 0.74/0.70/0.68     & 0.62/0.59/0.57     & -/-/-         & -/-/-        & -/-/-       \\ Swarm-SLAM~\cite{swarmslam}               & 4.62/4.17/3.90             & 6.50/5.27/4.39             & 5.56/4.72/4.15              & 2.69/2.48/2.34             & 8.53/7.59/7.10             & 5.61/5.04/4.72              & 1.61/1.33/1.09             & 1.98/1.48/0.94             & 1.80/1.41/1.02    \\
CP-SLAM~\cite{CP-SLAM}               & 6.21/5.56/5.27             & 5.67/5.37/4.67             & 5.73/5.26/4.77              & 1.45/1.43/1.39             & 2.48/2.32/2.27             & 1.85/1.68/1.73              & 0.62/0.47/0.30             & 1.28/1.17/1.37             & 0.91/0.78/0.80     \\
MNE-SLAM~\cite{mneslam}                 & \cellcolor{tabthird}1.21/1.09/1.07      & \cellcolor{tabthird}0.99/0.88/0.94     & \cellcolor{tabsecond}1.02/1.01/0.99     & \cellcolor{tabsecond}0.43/0.41/0.40      & \cellcolor{tabthird}0.74/0.72/0.70     & \cellcolor{tabsecond}0.59/0.58/0.55     & \cellcolor{tabsecond}0.43/0.41/0.40      & \cellcolor{tabsecond}0.54/0.51/0.50     & \cellcolor{tabsecond}0.48/0.46/0.45 \\
Ours                 & \cellcolor{tabsecond}1.18/1.06/1.04      & \cellcolor{tabsecond}0.97/0.87/0.93     & \cellcolor{tabfirst}1.00/1.00/0.96     & \cellcolor{tabfirst}0.41/0.39/0.38      & \cellcolor{tabsecond}0.71/0.70/0.68     & \cellcolor{tabfirst}0.57/0.56/0.53     & \cellcolor{tabfirst}0.42/0.40/0.38      & \cellcolor{tabfirst}0.52/0.49/0.48     & \cellcolor{tabfirst}0.46/0.44/0.43    \\ \bottomrule
\end{tabular}}}
  
\caption{Multi-agent tracking performance in the Apartment dataset~\cite{replica}. ATE RMSE($\downarrow$), Mean($\downarrow$) and Median($\downarrow$) are used as
evaluation metrics. Following the setting of \cite{CP-SLAM}, we quantitatively evaluated respective trajectories (part 1 and part 2) and average
results of the two agents. "-" indicates invalid results due to the
failure of CCM-SLAM. Our
method achieve SOTA performance compared with other existing methods. }
\label{tab:apartment}
   
\end{table*}

 \subsection{Multi-agent Communication }
 In our distributed multi-agent SLAM framework, when communication becomes available, robot $\alpha$ initiates the distributed loop closure detection process by sending the visual descriptors of its newly added keyframes to another robot $\beta$. For multi-agent communication, we use NetVLAD~\cite{netvlad} visual features to represent each keyframe. The overall information flow is illustrated in Fig.~\ref{fig:system}.
Loop closures are detected by computing the similarity between these visual descriptors.
Throughout all our experiments, we follow the same setting as Kimera-Multi~\cite{kimera-multi}: robots are constantly
in communication range, ensuring that inter-robot loop closures are triggered at the earliest possible opportunity. When a loop closure is detected, a signal is sent to the two corresponding agents, triggering information exchange and fusion. This includes transmitting the keyframe poses and network parameters of two agents, optimizing the keyframe pairs between the agents. By using the aligned poses and the keyframe depths, we can obtain the boundaries of the two agents' maps and compute the overlap region. The maps are then quickly fused using the distillation method. The different agents work simultaneously and merge submaps in real-time. Only the keyframe features, keyframe poses, and the network parameters for map construction are exchanged, and all computations performed locally on each agent. Compared to the traditional sparse point cloud map representation used in multi-agent systems, the neural implicit scene representation reduces the amount of information required for communication. Sparse point cloud maps tend to grow explosively as operation time and map scale increase.
\section{Experiments}
We evaluate the performance of our method on a wide range of publicly available datasets as well as real-world experiments. The evaluation covers diverse environments, including small-room scenes, multi-room layouts, long corridors, and large-scale outdoor scenes. We collect a new large-scale dataset, Dense SLAM datasets(DES), to further benchmark our system. Experimental results demonstrate that our method consistently outperforms existing single-agent and multi-agent SLAM methods in terms of surface reconstruction quality, pose estimation accuracy, and real-time performance.


\noindent\textbf{Datasets.} We evaluate MCN-SLAM on a variety of scenes from different datasets.
\begin{itemize}
    \item Replica Dataset~\cite{replica}.  8 small room scenes (\textbf{nearly}  $6.5m\times4.2m\times2.7m $ with 2000 images). We partitioned the dataset into two subsets, each corresponding to the trajectory of one agent, and ensured that there is overlap between the two subsets. We use this dataset to evaluate the reconstruction and localization accuracy in small-scale environments.
    \item ScanNet dataset ~\cite{scannet}. Real-world scenes with long sequences (more than 5000 images) and large-scale indoor scenarios. (\textbf{nearly} $7.5m\times6.6m\times3.5m $). We use this dataset for large-scale real-world indoor environments.
    \item Apartment dataset from~\cite{replica}. Multi-rooms scene (\textbf{nearly} $10.8m\times8.3m\times3.2m $ with more than 6000 images). We use this dataset for multi-room environments.
    \item TUM RGB-D dataset from~\cite{tum}, we use this classic SLAM dataset to evalute the camera tracking performance of our framework and compare with traditional SLAM framework.
    \item Our own dataset for Dense SLAM systems (DES dataset). This is the first real-world dataset for all kinds of neural SLAM systems with high-accuracy ground-truth for both camera trajectory and 3D reconstruction mesh. We collected single-agent and multi-agent datasets from various indoor environments, ranging from small room scenes (\textbf{nearly} $35m^2$)  to  large-scale outdoor scenes (\textbf{$>$276,000 $m^2$}), accumulating a total of 500,000 camera frames. The details are shown in Fig.~\ref{fig:dataset1},\ref{fig:dataset2}. We present detailed information on the dataset sequences and the specifications of the sensors. The dataset will be open-sourced on \href{https://dtc111111.github.io/DES-dataset/}{https://dtc111111.github.io/DES-dataset/}.
\end{itemize}

\begin{figure*}[h]
    \centering
    \includegraphics[width=\linewidth]{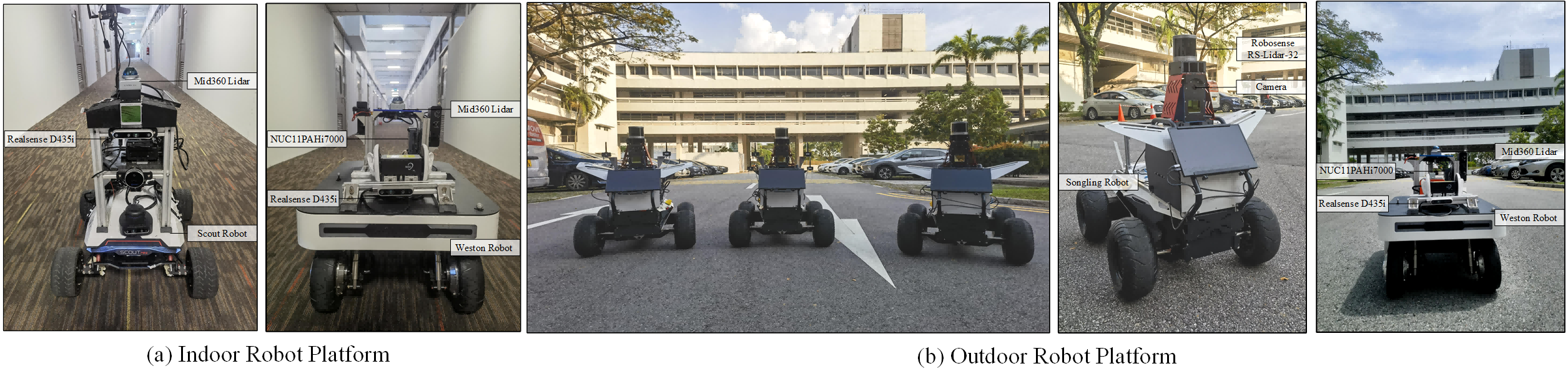}
    \caption{
    Visualization of the indoor and outdoor robot platform: Scout robot, Weston Robot, Songling Robot and camera and lidar sensors in our DES dataset. 
}
    \label{fig:robot}
\end{figure*}

\begin{figure}[h]
    \centering
    \includegraphics[width=0.9\linewidth]{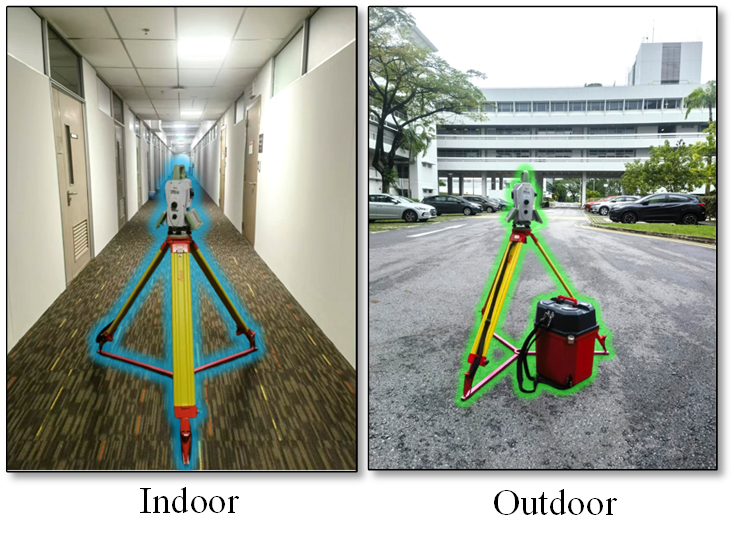}
       
    \caption{
    Visualization of the indoor and outdoor 3D Groundtruth collection platform Leica-rtc360
 in our DES dataset. 
}
    \label{fig:sensor}
      
\end{figure}

\begin{table}[]
\centering
\scalebox{0.85}{
\setlength{\tabcolsep}{1.2mm}{
\begin{tabular}{ll}
\toprule
Reconstruction Metrics        & Definition \\ \hline
Depth L1 [cm]         &$\frac{1}{N} \sum (\left|d_i-d_i^*\right|)/d_i$         \\
Accuracy [cm]        & $\sum_{p \in P}\left(\min _{q \in Q}\|p-q\|\right) /|P| $           \\
Completion [cm]      & $\sum_{q \in Q}\left(\min _{p \in P}\|p-q\|\right) /|Q|$          \\
Completion Ratio [$<5cm \%$] & $\sum_{q \in Q}\left(\min _{p \in P}\|p-q\|<0.05\right) /|Q|$         \\
 \bottomrule
\end{tabular}}}
  
\caption{Definitions of scene reconstruction metrics used for evaluation of surface reconstruction quality.}
\label{tab:metric}
   
\end{table}

\begin{figure*}[h]
    \centering
    \includegraphics[width=\linewidth]{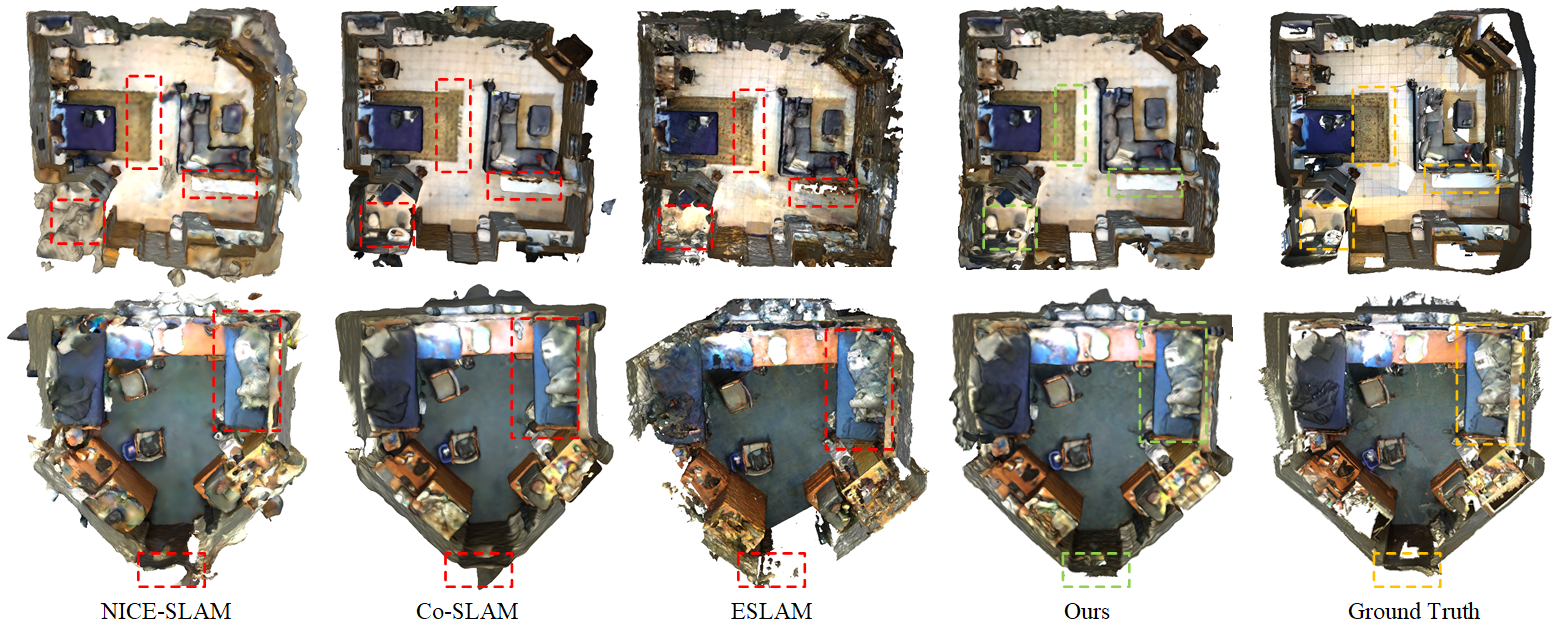}
       
    \caption{
    Two-agent experiments on the ScanNet dataset \cite{scannet} Scene 0000, 0233. We present Qualitative comparison of our proposed our method’s 3D surface reconstruction with existing NeRF-based dense visual SLAM methods, Co-SLAM \cite{coslam}, and ESLAM \cite{eslam}.  The region outlined on the image is marked in red to signify lower reconstruction accuracy, in green to signify higher accuracy, and in yellow to represent the groundtruth results. The experiment results demonstrates the effectiveness of the proposed scene representation method.}
    \label{fig:scannet}
      
\end{figure*}

\begin{figure*}[ht]
    \centering
    \includegraphics[width=\linewidth]{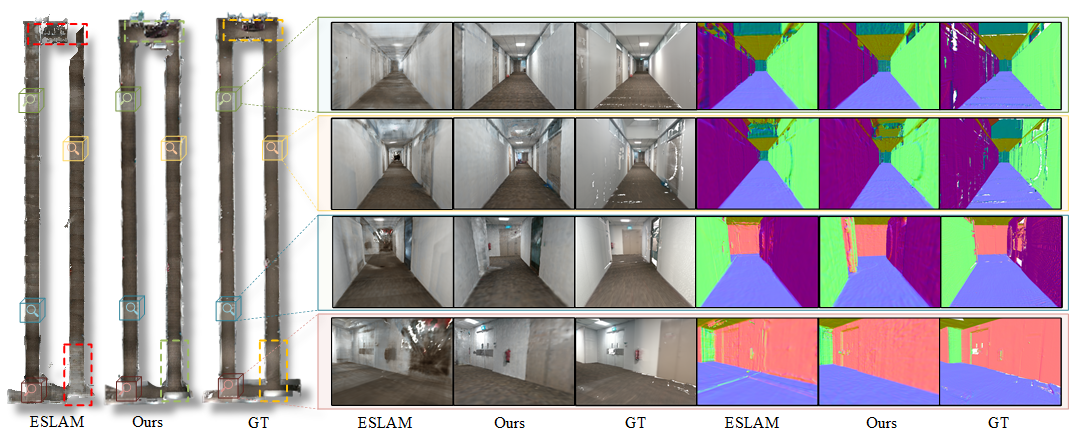}
       
    \caption{
     Multi-agent (four agents) experiments on our DES dataset. We present Qualitative results of our proposed method’s 3D surface reconstruction with existing SOTA method: ESLAM~\cite{eslam} on the long corridor sequence. The left side shows the fused mesh map reconstructed by four agents after collaborative SLAM. On the right is the zoomed-in view of the reconstructed mesh. The region outlined on the image is marked in red to signify lower reconstruction accuracy, in green to signify higher accuracy, and in yellow to represent the ground truth results. The right figure showcases the visualizations of the mesh and the corresponding normal maps.}
    \label{fig:indoor}
      
\end{figure*}

\noindent \textbf{Metrics.}
Following Neural-RGBD~\cite{neural_rgbd} and GO-Surf~\cite{gosurf}, we perform frustum and occlusion mesh culling that removes unobserved regions outside frustum and the noisy points within the
camera frustum but outside the target scene. Follow \cite{coslam}, We introduce a modification to the culling strategy used for the quantitative evaluation of the reconstruction accuracy, which leads to a fairer comparison.  We use the \textit{frustum+occlusion+virtul camera} culling method. This method simulates virtual camera views that cover the occluded regions. After mesh culling, we use Depth L1 (cm), Accuracy (cm), Completion (cm), and Completion ratio $(\%)$ to evaluate the reconstruction quality. The Depth L1 is defined as the average L1 difference between rendered GT depth and rendered depth. In Tab. \ref{tab:metric}, we present the 3D reconstruction metrics. We first
uniformly sample two point clouds $P$ and $Q$ from both GT and reconstructed meshes, with $|P| = |Q| = 200000$. Accuracy metric is defined as the average distance between a point on GT mesh to its nearest point on reconstructed mesh. The Completion metric is defined as the average distance between a point on reconstructed mesh to its nearest point on GT mesh. The Completion Ratio metric refers to the proportion of the overall ground truth (GT) where the average distance between a point on the reconstructed mesh and  its nearest point on the GT mesh is less than the threshold $t$. For indoor scenes, such as Replica~\cite{replica}, ScanNet~\cite{scannet}, and Apartment~\cite{replica} datasets, we set the threshold to 5 cm for completion ratio thresholds. 

For 2D metrics, we use  PSNR$\uparrow$, SSIM$\uparrow$, LPIPS$\downarrow$  metrics. We render RGB and depth images along with the trajectory of camera. 
For the evaluation of camera tracking, we adopt ATE RMSE(cm), Mean(cm), Median(cm). For the camera trajectories generated by CCM-SLAM~\cite{ccmslam}, we align them
with the Ground Truth camera trajectory using Sim(3) Umeyama alignment in the EVO tool. As for the camera trajectories produced by other methods, we align them with the Ground Truth camera trajectory by aligning the origin. Trajectory alignment is crucial for proper drift and loop closure evaluation. To be specific, after aligning the initial poses, we calculate the Absolute Trajectory Error (ATE) for each pose and compute the RMSE, Mean, and Median values.

\begin{table}[]
\centering
\scalebox{0.88}{
\setlength{\tabcolsep}{1.2mm}{
\begin{tabular}{lccccccccc}
\toprule
Methods       & 0000   & 0059   & 0106  & 0169  & 0181  & 0207  & 0054   & 0233  & Avg. \\ \midrule
\multicolumn{10}{l}{\cellcolor[HTML]{EEEEEE}{\textit{Single-Agent Methods}}} \\
NICE-SLAM~\cite{niceslam}   & 12.0 & 14.0 & 7.9  & 10.9 & 13.4 & 6.5  & 20.9 & 9.0  & 13.0 \\
ESLAM~\cite{eslam}     & 7.3  & 8.5  & 7.5  & 7.5  & 9.0  & \cellcolor{tabsecond}6.2  & 36.3 & 4.6  & 10.6 \\
Co-SLAM~\cite{coslam}      & 7.1  & 11.1 & 9.4  & \cellcolor{tabfirst}7.1  & 11.8 & 7.1  & 14.8   & -    & -    \\
Point-SLAM~\cite{pointslam}   & 10.2 & 7.8  & 8.7  & 22.2 & 14.8 & 9.5  & 28.0 & 6.1  & 14.3 \\
GO-SLAM~\cite{goslam}       & 5.4  &\cellcolor{tabsecond}7.5  & \cellcolor{tabsecond}7.0  & 7.7  & \cellcolor{tabthird}6.8  & 6.9  & \cellcolor{tabsecond}8.8  & 4.8  &6.9  \\
PLGSLAM~\cite{plgslam}   & 7.3  & 8.2  & 7.4  & 7.5  & 9.2  & \cellcolor{tabfirst}5.8  & 30.8 & \cellcolor{tabfirst}4.2  & 9.9  \\
Loopy-SLAM~\cite{loopyslam}   & \cellcolor{tabsecond}5.2  &7.9  & 8.5  & 7.7  & 10.6 & 7.9  & 14.5  & 5.2  & \cellcolor{tabthird}7.7  \\
\multicolumn{10}{l}{\cellcolor[HTML]{EEEEEE}{\textit{Multi-Agent Methods}}} \\
Swarm-SLAM~\cite{swarmslam}    & 10.7 & 8.7  & 9.5  & 12.5 & 13.8 & 8.4  & 14.5 & 6.5  & 10.5 \\
CP-SLAM~\cite{CP-SLAM} & 7.6  & 8.9  & 8.8  & 8.1  & 10.8  & 8.3  & 20.1 & 5.6  & 9.7 \\
MNE-SLAM~\cite{mneslam}       & \cellcolor{tabsecond}5.0  & \cellcolor{tabsecond}7.3  & \cellcolor{tabsecond}6.6  & \cellcolor{tabthird}7.3  & \cellcolor{tabsecond}6.4  & 6.4  & \cellcolor{tabsecond}8.5  & \cellcolor{tabthird}4.3  &\cellcolor{tabfirst}6.6 \\
Ours       & \cellcolor{tabfirst}4.7  & \cellcolor{tabfirst}7.0  & \cellcolor{tabfirst}6.5  & \cellcolor{tabsecond}7.2  & \cellcolor{tabfirst}6.2  & \cellcolor{tabsecond}6.2  & \cellcolor{tabfirst}8.3  & \cellcolor{tabfirst}4.2  &\cellcolor{tabfirst}6.4 \\ \bottomrule
\end{tabular}}}
 
\caption{Multi-agent (two agents) camera tracking performance on ScanNet dataset~\cite{scannet}. We use ATE RMSE(cm) as the metric. We select 8 sequences follow the setting from previous methods.}
  
\label{tab:scannet}
\end{table}

\noindent \textbf{Baseline Methods.} We compare our multi-agent method with the current state-of-the-art baseline methods, including traditional methods, single-agent NeRF-SLAM methods, 3DGS-SLAM methods, and multi-agent NeRF-SLAM methods.
For the traditional methods, we choose ORB-SLAM3~\cite{orbslam3}, BAD-SLAM~\cite{BADslam}, Kintinuous~\cite{kintinuous}, ElasticFusion~\cite{elasticfusion}, BundleFusion~\cite{Bundle-fusion}. For single-agent nerf-based methods, we choose DI-Fusion~\cite{difusion}, NICE-SLAM~\cite{niceslam}, Vox-Fusion~\cite{voxfusion}, MIPS-Fusion~\cite{mipsfusion}, Point-SLAM~\cite{pointslam}, ESLAM~\cite{eslam}, Co-SLAM~\cite{coslam}, PLGSLAM~\cite{plgslam}, GO-SLAM~\cite{goslam}, Loopy-SLAM~\cite{loopyslam}. We also choose a 3DGS-based SLAM
 method, Photo-SLAM~\cite{photoslam}. For multi-agents methods, we choose CCM-SLAM~\cite{ccmslam}, Swarm-SLAM~\cite{swarmslam}, and CP-SLAM~\cite{CP-SLAM}. Since recent multi-agent methods (Kimera-Multi~\cite{kimera-multi}, $D^2$SLAM~\cite{D2slam}) are based on tightly-coupled camera-IMU systems and are not designed for dense mapping, we exclude these methods from our comparison.
 
\noindent \textbf{Implementation.} We employ feature planes
with a resolution of 24 cm for coarse tri-planes. we use L = 16 level HashGrid with from Rmin = 16 to Rmax, where we use max voxel size 2cm for determining Rmax.  All feature planes have 32 channels, resulting in a 64-channel concatenated feature input for the decoders. The decoders are two-layer MLPs with 32 channels in the hidden layer. 
The dimension of the geometric feature $\mathbf{z}$ is 15. ReLU activation function is used for the hidden layer, and Tanh and Sigmoid are respectively used for
the output layers of TSDF and raw colors.
We use 16 bins
for One-Blob encoding of each dimension. 
For Replica \cite{replica} dataset,
we sample $K$ = 32 points for regular sampling and
$K_{surface}$ = 11 points for depth-guided sampling on each ray. We use 200 iterations
for first frame mapping. We perform 10 optimization iterations for mapping and randomly select 4000 rays for
each iteration. And for ScanNet~\cite{scannet} dataset, we set $K$ = 96 and $K_{surface}$ = 21. Also, we perform 30 optimization iterations for  mapping 
in ScanNet scenes. For the scenes in Apartment
dataset \cite{niceslam}, we similarly set $K$ = 96 and $K_{surface}$ = 21. For this dataset, We perform 30 optimization iterations for
mapping, and
we randomly sample 5000 rays for each iteration. We use the RAFT~\cite{raft} feature to select proporiate keyframe for jointly optimizing the hybrid scene representation network, and camera poses of
the selected keyframes. We add a keyframe when the average flow  is greater than or equal to 4 pixels. We use Adam \cite{adam} for optimizing all learnable parameters of our method. Once all input frames are processed, and for evaluation purposes, we build a TSDF volume for each scene and use the marching cubes algorithm \cite{marching} to obtain 3D meshes.

\begin{figure*}[h]
    \centering
    \includegraphics[width=\linewidth]{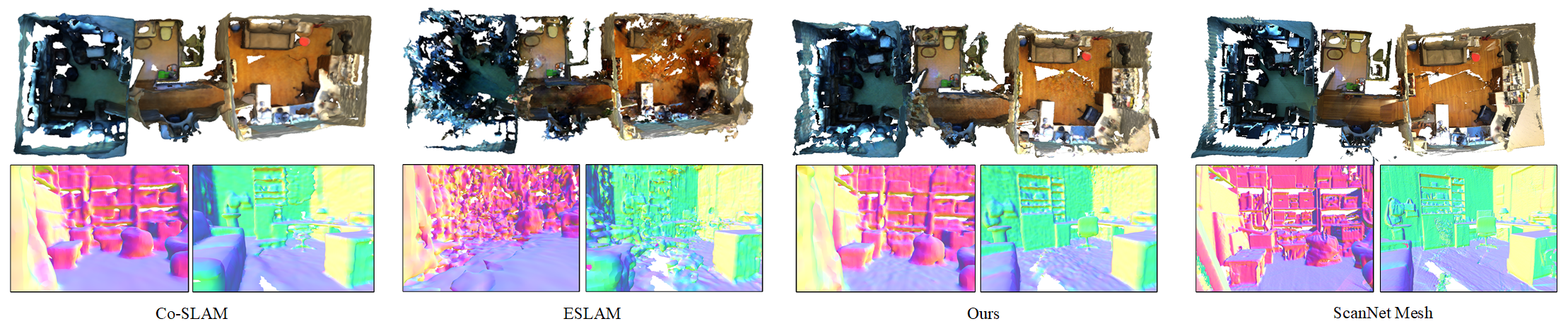}
       
    \caption{
    Multi-agent (Two-agents) experiments of our proposed method’s surface reconstruction with existing SOTA method: ESLAM \cite{eslam}, Co-SLAM~\cite{coslam} on the ScanNet dataset~\cite{scannet} Scene 0054. The top shows the reconstructed mesh, while the bottom displays zoomed-in normal images. 
}
   
    \label{fig:scannet2}
\end{figure*}

\begin{figure}[t]
    \centering
    \includegraphics[width=\linewidth]{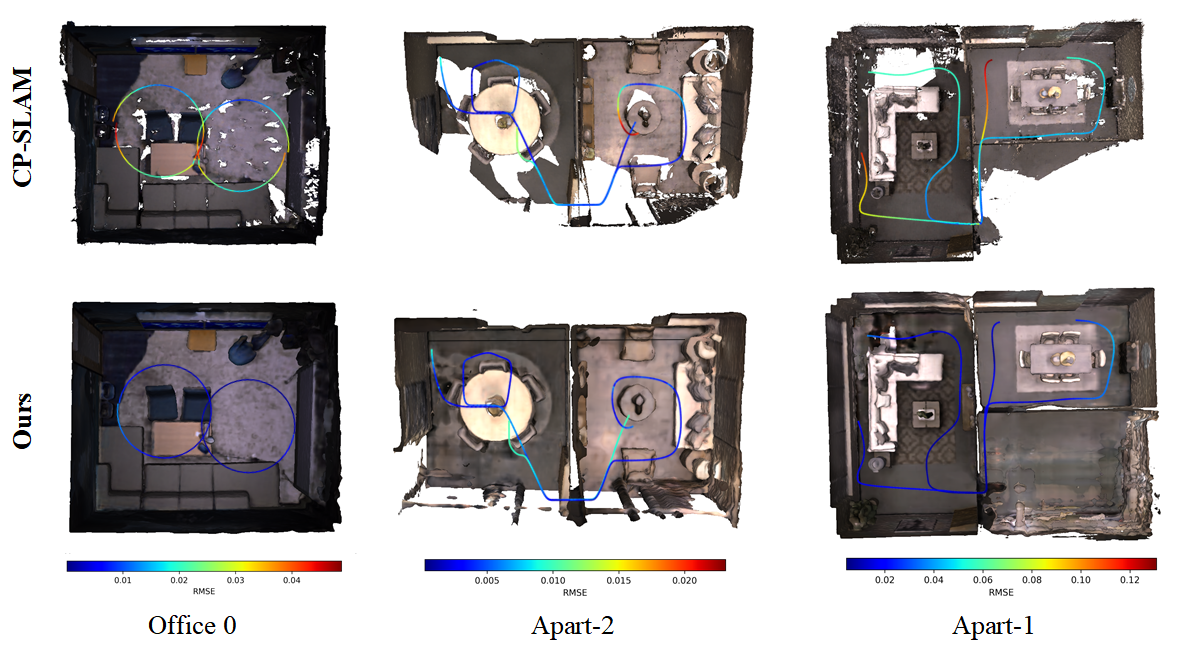}
       
    \caption{
   Multi-agent (Two agents) experiments of our proposed method’s surface reconstruction with existing SOTA method CP-SLAM~\cite{CP-SLAM} on the Replica dataset and Apartment dataset~\cite{replica}. A colorbar indicating trajectory RMSE is shown below each figure, where red represents higher error and blue indicates lower error. }
   
  
    \label{fig:apartment}
\end{figure}

\begin{figure*}[t]
    \centering
    \includegraphics[width=0.95\linewidth]{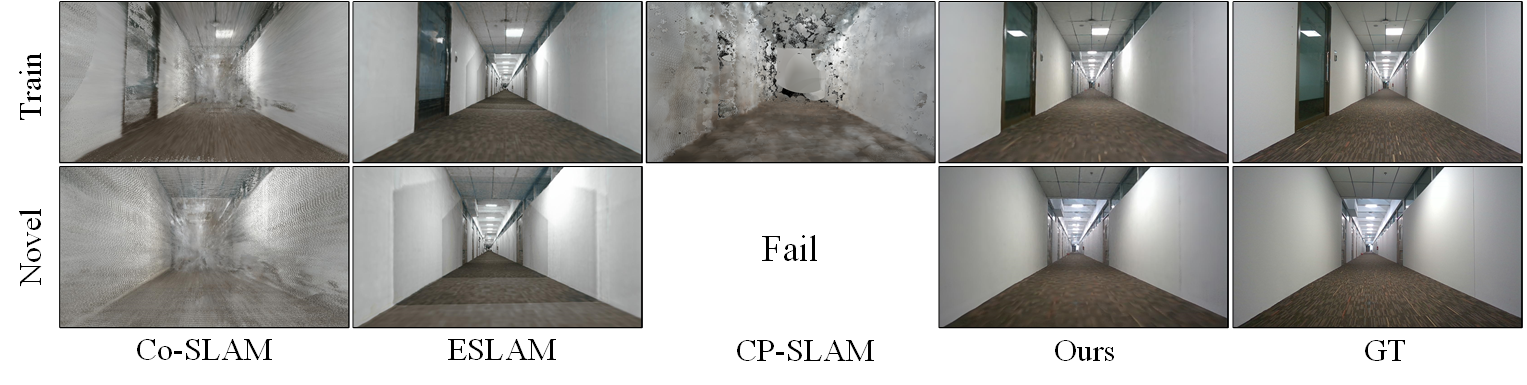}
      
    \caption{
    Visualization of rendering performance of four agents experiments on DES dataset (indoor corridor sequence). We present the rendered training views and novel views. We compare our proposed method’s rendering performance with existing SOTA methods on our DES dataset. 
}
  
    \label{fig:render}
\end{figure*}

\begin{table*}[]

\centering
\scalebox{0.73}{
\setlength{\tabcolsep}{0.4mm}{
\begin{tabular}{lcccccccccccccccccccccccc}
\hline
\multirow{3}{*}{Methods} & \multicolumn{6}{c}{Agent1}                                                                                                                                                                                                        & \multicolumn{6}{c}{Agent2}                                                                                                                                                                                                        & \multicolumn{6}{c}{Agent3}                                                                                                                                                                                                        & \multicolumn{6}{c}{Agent4}                                                                                \\
                         & \multicolumn{3}{c}{Training View}                                                                               & \multicolumn{3}{c}{Novel View}                                                                                  & \multicolumn{3}{c}{Training View}                                                                               & \multicolumn{3}{c}{Novel View}                                                                                  & \multicolumn{3}{c}{Training View}                                                                               & \multicolumn{3}{c}{Novel View}                                                                                  & \multicolumn{3}{c}{Training View}                   & \multicolumn{3}{c}{Novel View}                      \\
                         & \multicolumn{1}{l}{PNSR$\uparrow$} & \multicolumn{1}{l}{SSIM$\uparrow$} & \multicolumn{1}{l}{LPIPS$\downarrow$} & \multicolumn{1}{l}{PNSR$\uparrow$} & \multicolumn{1}{l}{SSIM$\uparrow$} & \multicolumn{1}{l}{LPIPS$\downarrow$} & \multicolumn{1}{l}{PNSR$\uparrow$} & \multicolumn{1}{l}{SSIM$\uparrow$} & \multicolumn{1}{l}{LPIPS$\downarrow$} & \multicolumn{1}{l}{PNSR$\uparrow$} & \multicolumn{1}{l}{SSIM$\uparrow$} & \multicolumn{1}{l}{LPIPS$\downarrow$} & \multicolumn{1}{l}{PNSR$\uparrow$} & \multicolumn{1}{l}{SSIM$\uparrow$} & \multicolumn{1}{l}{LPIPS$\downarrow$} & \multicolumn{1}{l}{PNSR$\uparrow$} & \multicolumn{1}{l}{SSIM$\uparrow$} & \multicolumn{1}{l}{LPIPS$\downarrow$} & PNSR$\uparrow$ & SSIM$\uparrow$ & LPIPS$\downarrow$ & PNSR$\uparrow$ & SSIM$\uparrow$ & LPIPS$\downarrow$ \\ \hline
                         \multicolumn{25}{l}{\cellcolor[HTML]{EEEEEE}{\textit{Single-Agent Methods}}} \\
Co-SLAM~\cite{coslam}                  & 14.7                               & 0.49                               & 0.79                                  & 14.2                               & 0.39                               & 0.89                                  & 14.9                               & 0.50                               & 0.78                                  & 14.2                               & 0.39                               & 0.87                                  & 14.3                               & 0.47                               & 0.80                                  & 13.9                               & 0.39                               & 0.89                                  & 15.3           & 0.49           & 0.79              & 14.5           & 0.41           & 0.88              \\
ESLAM~\cite{eslam}                    & 25.8                               & 0.84                               & 0.51                                  & 19.8                               & 0.79                               & 0.50                                  & 25.9                               & 0.85                               & 0.51                                  & 19.8                               & 0.80                               & 0.49                                  & 25.4                               & 0.82                               & 0.52                                  & 19.3                               & 0.79                               & 0.50                                  & 26.4           & 0.84           & 0.51              & 20.2           & 0.80           & 0.49              \\
Point-SLAM~\cite{pointslam}               & 26.5                               & \cellcolor{tabthird}0.86                               & 0.49                                  & \cellcolor{tabthird}25.9                               & \cellcolor{tabthird}0.85                               & \cellcolor{tabthird}0.43                                  & \cellcolor{tabthird}26.8                               & \cellcolor{tabthird}0.87                               & 0.48                                  & \cellcolor{tabthird}25.9                               & \cellcolor{tabthird}0.85                               & \cellcolor{tabthird}0.43                                  & 26.7                               & 0.84                               & 0.48                                  & \cellcolor{tabthird}25.4                               & \cellcolor{tabthird}0.84                               & \cellcolor{tabthird}0.43                                  & 26.9           & \cellcolor{tabthird}0.86           & 0.49              & \cellcolor{tabthird}26.1           & \cellcolor{tabthird}0.86           & \cellcolor{tabthird}0.42              \\
PLGSLAM~\cite{plgslam}                  & 25.9                               & 0.85                               & 0.49                                  & 20.3                               & 0.81                               & 0.49                                  & 26.1                               & 0.86                               & 0.48                                  & 20.3                               & 0.81                               & 0.48                                  & 25.5                               & 0.84                               & 0.51                                  & 19.9                               & 0.80                               & 0.49                                  & 26.2           & 0.85           & 0.49              & 20.8           & 0.82           & 0.48              \\
Photo-SLAM~\cite{photoslam}               & \cellcolor{tabthird}28.3                               & \cellcolor{tabthird}0.86                               & \cellcolor{tabthird}0.43                                  & 25.7                               & 0.84                               & 0.44                                  & 28.7                               & 0.86                               & \cellcolor{tabthird}0.43                                  & 25.7                               & \cellcolor{tabthird}0.85                               & 0.44                                  & \cellcolor{tabthird}28.0                               & \cellcolor{tabthird}0.85                               & \cellcolor{tabthird}0.45                                  & 25.2                               & \cellcolor{tabthird}0.84                               & 0.45                                  & \cellcolor{tabthird}28.7           & \cellcolor{tabthird}0.86           & \cellcolor{tabthird}0.43              & 25.9           & 0.85           & 0.43              \\
\multicolumn{25}{l}{\cellcolor[HTML]{EEEEEE}{\textit{Multi-Agent Methods}}} \\
CP-SLAM~\cite{CP-SLAM}                  & 14.3                               & 0.47                               & 0.82                                  & -                                  & -                                  & -                                     & 14.7                               & 0.47                               & 0.80                                  & -                                  & -                                  & -                                     & 14.5                               & 0.46                               & 0.84                                  & -                                  & -                                  & -                                     & 14.9           & 0.47           & 0.82              & -              & -              & -                 \\
MNE-SLAM~\cite{mneslam}                     & \cellcolor{tabsecond}29.5                               & \cellcolor{tabsecond}0.88                               & \cellcolor{tabsecond}0.41                                  & \cellcolor{tabsecond}27.8                               & \cellcolor{tabsecond}0.87                               & \cellcolor{tabsecond}0.42                                  & \cellcolor{tabsecond}30.0                               & \cellcolor{tabsecond}0.89                               & \cellcolor{tabsecond}0.39                                  & \cellcolor{tabsecond}28.2                               & \cellcolor{tabsecond}0.88                               & \cellcolor{tabsecond}0.41                                  & \cellcolor{tabsecond}29.1                               & \cellcolor{tabsecond}0.87                               & \cellcolor{tabsecond}0.40                                  & \cellcolor{tabsecond}27.2                               & \cellcolor{tabsecond}0.81                               & \cellcolor{tabsecond}0.43                                  & \cellcolor{tabsecond}30.1           & \cellcolor{tabsecond}0.89           & \cellcolor{tabsecond}0.39              & \cellcolor{tabsecond}28.3           & \cellcolor{tabsecond}0.89           & \cellcolor{tabsecond}0.39              \\
Ours                     & \cellcolor{tabsecond}30.2                               & \cellcolor{tabfirst}0.92                               & \cellcolor{tabfirst}0.38                                  & \cellcolor{tabfirst}28.5                               & \cellcolor{tabfirst}0.90                               & \cellcolor{tabfirst}0.39                                  & \cellcolor{tabfirst}30.7                               & \cellcolor{tabfirst}0.92                               & \cellcolor{tabfirst}0.37                                  & \cellcolor{tabfirst}28.9                               & \cellcolor{tabfirst}0.90                               & \cellcolor{tabfirst}0.39                                  & \cellcolor{tabfirst}29.5                               & \cellcolor{tabfirst}0.89                               & \cellcolor{tabfirst}0.40                                  & \cellcolor{tabfirst}27.9                               & \cellcolor{tabfirst}0.88                               & \cellcolor{tabfirst}0.41                                  & \cellcolor{tabfirst}30.8           & \cellcolor{tabfirst}0.91           & \cellcolor{tabfirst}0.39              & \cellcolor{tabfirst}28.9           & \cellcolor{tabfirst}0.90           & \cellcolor{tabfirst}0.38              \\ \bottomrule
\end{tabular}}}
 
\caption{Quantitative rendering results of multi-agent (four agents) experiments. We compare our proposed framework with the existing SLAM systems on DES dataset (long corridor sequence) with projected depth images.}
  
\label{tab:renderview}
\end{table*}

\begin{table*}[]
\centering
\scalebox{0.82}{
\setlength{\tabcolsep}{0.5mm}{
\begin{tabular}{lccccccccccccccc}
\toprule
\multirow{2}{*}{Methods} & \multicolumn{3}{c}{Agent1}                                      & \multicolumn{3}{c}{Agent2}                                      & \multicolumn{3}{c}{Agent3}                                      & \multicolumn{3}{c}{Agent4}                                      & \multicolumn{3}{c}{Global}                                      \\
                         & Acc.$\downarrow$ & Comp.$\downarrow$ & Comp.Ratio(\%)$\uparrow$ & Acc.$\downarrow$ & Comp.$\downarrow$ & Comp.Ratio(\%)$\uparrow$ & Acc.$\downarrow$ & Comp.$\downarrow$ & Comp.Ratio(\%)$\uparrow$ & Acc.$\downarrow$ & Comp.$\downarrow$ & Comp.Ratio(\%)$\uparrow$ & Acc.$\downarrow$ & Comp.$\downarrow$ & Comp.Ratio(\%)$\uparrow$ \\ \bottomrule
                         \multicolumn{16}{l}{\cellcolor[HTML]{EEEEEE}{\textit{Single-Agent Methods}}} \\
Co-SLAM~\cite{coslam}                 & 47.71 & 95.85 & 27.47
& 93.74 & 144.39 & 6.94
& 77.90 & 76.31 & 26.62
& 245.27 & 195.41 & 1.47                      & -                & -                 & -                        \\
ESLAM~\cite{eslam}           & 37.66 & 33.94 & 54.31
& 87.60 & 31.43 & 41.87
& 89.73 & 14.94 & \cellcolor{tabthird}65.84
& 102.13 & 46.44 & 43.87                    & 73.78            & 17.60             & 58.49                    \\
GO-SLAM~\cite{goslam}                  & 52.74          & 48.47           & 30.48                    & 94.39          & 44.93           & 31.49                    & 101.37          & 29.31           & 40.13                    & 142.59          & 89.33           & 30.15                    & 109.12          & 133.15           & 10.53                    \\
Point-SLAM~\cite{pointslam}              & 39.45 & 34.77 & 34.15
& 91.61 & 37.27 & 35.87
& 93.49 & 21.12 & 44.26
& 139.51 & 52.34 & 26.15                    & -            & -             & -                     \\
PLGSLAM~\cite{plgslam}                 & 37.29 & \cellcolor{tabthird}33.47 & \cellcolor{tabthird}55.19
& \cellcolor{tabthird}84.14 & 30.47 & 43.49
& \cellcolor{tabthird}85.19 & \cellcolor{tabthird}14.81 & \cellcolor{tabsecond}67.15
& 99.19 & \cellcolor{tabthird}42.16 & \cellcolor{tabthird}46.91                    & \cellcolor{tabthird}70.42             & \cellcolor{tabthird}15.49             & \cellcolor{tabthird}59.37                    \\
Photo-SLAM~\cite{photoslam}               & 50.15          & 43.29           & 34.17                    & 89.11          & \cellcolor{tabthird}30.25           & \cellcolor{tabthird}45.81                    & 87.14          & 27.56           & 47.39                    & \cellcolor{tabthird}98.59          & 63.49           & 31.78                              & 60.61            & 139.89            & 10.75                    \\
Loopy-SLAM~\cite{loopyslam}               & \cellcolor{tabthird}37.25 & 41.14 & 36.13
& 88.65 & 45.29 & 28.85
& 91.26 & 20.93 & 46.15
& 128.28 & 51.81 & 29.05                   & -            & -             & -                          \\ \multicolumn{16}{l}{\cellcolor[HTML]{EEEEEE}{\textit{Multi-Agent Methods}}} \\
CP-SLAM~\cite{CP-SLAM}                  & 44.01 & 92.09 & 29.79
& 91.43 & 142.08 & 9.14
& 73.74 & 73.36 & 24.75
& 242.11 & 164.46 & 4.18                      & -                & -                 & -                        \\
MNE-SLAM~\cite{mneslam}                     & \cellcolor{tabsecond}37.21 & \cellcolor{tabsecond}33.45 & \cellcolor{tabsecond}55.29
& \cellcolor{tabsecond}62.23 & \cellcolor{tabsecond}14.22 & \cellcolor{tabsecond}68.03
& \cellcolor{tabsecond}46.09 & \cellcolor{tabthird}14.83 & 64.76
& \cellcolor{tabsecond}68.95 & \cellcolor{tabsecond}13.34 & \cellcolor{tabsecond}75.71                  & \cellcolor{tabsecond}43.96            & \cellcolor{tabsecond}14.69             & \cellcolor{tabsecond}64.00                    \\
Ours                     & \cellcolor{tabfirst}37.03 & \cellcolor{tabfirst}33.41 & \cellcolor{tabfirst}55.85
& \cellcolor{tabfirst}62.09 & \cellcolor{tabfirst}14.16 & \cellcolor{tabfirst}68.58
& \cellcolor{tabfirst}46.04 & \cellcolor{tabsecond}14.79 & \cellcolor{tabfirst}67.37
& \cellcolor{tabfirst}68.74 & \cellcolor{tabfirst}13.29 & \cellcolor{tabfirst}75.85                  & \cellcolor{tabfirst}43.89            & \cellcolor{tabfirst}14.58             & \cellcolor{tabfirst}64.43                    \\  \bottomrule
\end{tabular}}}
 
\caption{Multi-agents (four agents) scene reconstruction performance on DES dataset (Long corridor scene) with sensor depth images. ``-" indicates invalid results due to the failure or inability of these methods. }
  
\label{tab:indoor}
\end{table*}

\subsection{Dataset Collection}
 We introduce a large-scale Neural SLAM dataset that spans a diverse range of scenarios, from small indoor rooms to expansive outdoor environments. The dataset encompasses both single-agent and multi-agent settings, providing high-quality 3D mesh ground truth and continuous-time trajectory ground truth. To the best of our knowledge, this is the first large-scale dataset for the dense SLAM to offer both 3D map ground truth and trajectory ground truth. We believe this dataset has the potential to significantly advance research and innovation within the SLAM community.

\noindent\textbf{Sensor Suit and Robot Platform} In Fig. ~\ref{fig:robot}, we present the four robotic platforms used for dataset collection, along with detailed sensor information, such as cameras and LiDAR. We use Scout robot and Weston Robot to collect data in indoor scenes. As for outdoor scenes, we use three Songling robot and a weston robot to collect data.

\noindent\textbf{Calibration} The calibration process involves finding the camera intrinsics and the extrinsics of all sensors.
First, the Kalibr toolbox is used to find the intrinsic parameters of the D455 cameras. These intrinsic parameters are then used to find the extrinsic of the camera setup on each D455 w.r.t. each IMU available. Finally, we perform a pose-graph optimization over the
graph to obtain the final extrinsic of all sensors and summarize them in the calibration report.

\noindent\textbf{3D Groundtruth Map}
3D survey-grade mapping of the campuses was done by using various industrial laser scanners. The 3D groundtruth map  collection platform is shown in Fig.~\ref{fig:sensor}. We visualize the
collection platform Leica-rtc360 in our DES dataset of the indoor and outdoor 3D Groundtruth. In essence, It acquires point cloud data from discrete, static locations by means of ground-based, high-resolution 3D laser scanners. These individual scans are then combined to produce a complete 3D point cloud via co-visible landmarks (both man-made and natural) and propriety global scan matching
software. point cloud data consists of 3D point coordinates and may also include intensity or RGB channels. Scanning
each section took between several days, requiring at least two field operators.
Upon completion of the scanning process, all individual scans were registered in the manufacturer’s software. Registration of individual scans in each of the sections was performed solely on the basis of the artificial sphere targets
recorded in the scans. Cloud-to-cloud registration was then used to align the sections and create the final point cloud of the entire area. Target-based statistics showed a maximum distance error of less than 10 mm for all sphere pairings that were used in the registration optimization process. Building walls and corners were spot-checked for overlap of scans. A total of
10 reference measurements were made with a total station to verify the global accuracy of the point cloud. The results
showed a maximum distance error of less than 5 cm over than entire length of the campus, which is below the resolution of the point cloud. We get the 3D mesh
use the groundtruth point cloud with Poisson Surface Reconstruction and TSDF fusion. Due to space constraints, we have included the dataset visualizations in the supplementary material, as shown in Fig. ~\ref{fig:dataset1}. We present several viewpoints from the dataset, with the
bottom-right corner showing the 3D ground truth for some scenes. Additionally, following the tools provided in iMAP~\cite{imap}, users can also generate custom camera trajectories along with corresponding RGB and depth images using the 3D ground truth by themselves.

\begin{table}[]
\centering
\scalebox{0.76}{
\setlength{\tabcolsep}{0.4mm}{
\begin{tabular}{lcccccc}
\toprule
   \multirow{2}{*}{Methods}  & fr1/desk &  fr1/desk2 & fr1/room & fr2/xyz & fr3/office & Avg. 
   \\ 
   & RMSE/Mean &RMSE/Mean &RMSE/Mean &RMSE/Mean
   &RMSE/Mean &RMSE/Mean \\
\midrule
\multicolumn{7}{l}{\cellcolor[HTML]{EEEEEE}{\textit{Traditional Methods}}} \\
BAD-SLAM~\cite{BADslam}        &     1.75/1.68 &  N/A &   N/A &     1.14/1.09 & 1.74/1.70 &  N/A\\
Kintinuous~\cite{kintinuous}  & 3.74/3.57  & 7.15/7.03 &   7.54/7.38  & 2.95/2.81 & 3.03/2.90 &      4.88/4.69\\
ORBSLAM3~\cite{orbslam3}  &   \cellcolor{tabthird}1.64/1.57 &      \cellcolor{tabfirst}2.24/2.13 &   \cellcolor{tabthird}4.65/4.45 &    \cellcolor{tabfirst}0.44/0.40 &   \cellcolor{tabfirst}1.06/1.02 &     \cellcolor{tabfirst}2.02/1.92\\
ElasticFusion~\cite{elasticfusion}  &  2.53/2.41 & 6.83/6.41 & 21.49/19.96 &  1.17/1.04 & 2.52/2.34 & 6.91/6.65\\
BundleFusion~\cite{Bundle-fusion}  &  1.65/1.57 &  N/A &   N/A & 1.15/1.04 & 2.25/2.13 &  N/A\\ \hdashline
\multicolumn{7}{l}{\cellcolor[HTML]{EEEEEE}{\textit{Neural-based Methods}}} \\
DI-Fusion~\cite{difusion}          & 4.45/4.21  &  N/A & N/A & 2.04/1.98 & 5.88/5.67 &  N/A\\
NICE-SLAM~\cite{niceslam}      & 4.26/4.01 &  4.99/4.68  &34.49/33.12  &6.19/5.86  &3.87/3.67 &10.76/10.19 \\
Vox-Fusion~\cite{voxfusion}       & 3.52/3.41 & 6.00/5.47  &  19.53/18.37 & 1.49/1.27&  26.01/25.31 &   11.31/10.84 \\ 
MIPS-Fusion~\cite{mipsfusion}       & 3.04/2.71  & N/A  & N/A & 1.44/1.31 &  4.63/4.31 &  N/A \\ 
Point-SLAM~\cite{pointslam}  & 4.34/4.09 & 4.54/4.28  &  30.92/29.45  &1.31/1.18  &3.48/3.27 &  8.92/7.91 \\
ESLAM~\cite{eslam}          & 2.47/2.35 &     3.69/3.46  & 29.73/28.31 & 1.11/1.02 &  2.42/2.25 &  7.89/7.03 \\ 
Co-SLAM~\cite{coslam}       & 2.40/2.25 & N/A  & N/A & 1.74/1.64 &  2.45/2.27 &  N/A \\ 
PLGSLAM~\cite{plgslam}          & 2.45/2.31 &     3.67/3.34  & 29.61/28.08 & 1.10/1.06 &  2.40/2.24 &  7.85/7.47 \\
GO-SLAM~\cite{goslam}          &  \cellcolor{tabsecond}1.52/1.43  & \cellcolor{tabthird}2.78/2.62  &  \cellcolor{tabsecond}4.64/4.41 &  \cellcolor{tabthird}0.69/0.65 &   \cellcolor{tabthird}1.39/1.31 & \cellcolor{tabthird}2.20/2.08  \\ 
Loopy-SLAM~\cite{loopyslam}             & 3.79/3.45  &   3.38/3.15  &  7.03/6.29  & 1.62/1.51 & 3.41/3.24 &  3.85/3.52  \\
Ours            &  \cellcolor{tabfirst}1.42/1.33  & \cellcolor{tabsecond}2.45/2.27  &  \cellcolor{tabthird}4.50/4.23 &  \cellcolor{tabfirst}0.62/0.58 &   \cellcolor{tabsecond}1.31/1.20 & \cellcolor{tabsecond}2.06/1.92 \\ \bottomrule
\end{tabular}}}
 
\caption{Single agent experiments on camera tracking performance on TUM-RGBD~\cite{tum} (ATE RMSE $\downarrow$(cm)/Mean(cm)$\downarrow$).  shows competitive performance on a variety of scenes.}
  
\label{tab:tum}
\end{table}

\begin{table}[]
\centering
\scalebox{0.83}{
\setlength{\tabcolsep}{0.5mm}{
\begin{tabular}{lcccccccccc}
\toprule
\multirow{2}{*}{Methods} & \multicolumn{2}{c}{Agent1}            & \multicolumn{2}{c}{Agent2}            & \multicolumn{2}{c}{Agent3}            & \multicolumn{2}{c}{Agent4}            & \multicolumn{2}{c}{Global}            \\
                         & Mean & RMSE  & Mean  & RMSE  & Mean  & RMSE  & Mean  & RMSE  & Mean  & RMSE \\
                         \multicolumn{11}{l}{\cellcolor[HTML]{EEEEEE}{\textit{Single-Agent Methods}}} \\
ORB-SLAM3~\cite{orbslam3}                & 0.43              & 0.47              & 0.31              & 0.35              & 0.29              & 0.32              & 0.31              & 0.36            & 0.76              & 0.80              \\
Co-SLAM~\cite{coslam}                  & 10.22             &  12.05            & 12.46            & 14.99             & 9.29              & 11.97            & 9.33              & 10.47             & 16.84             & 18.80             \\
ESLAM~\cite{eslam}                    & 1.22              &  1.33              & 0.45              & 0.51              & 0.31               & 0.39              & 0.68               & 0.81             & 1.10              & 1.23              \\
GO-SLAM~\cite{goslam}                  & \cellcolor{tabthird}0.13              & \cellcolor{tabthird}0.17              & 0.43              & 0.48              & \cellcolor{tabthird}0.21              & \cellcolor{tabthird}0.24              & 0.33              & 0.37              & 3.76              & 4.36              \\
Point-SLAM~\cite{pointslam}               & 2.11              & 2.25              & 1.14              & 1.19              & 0.87              & 0.93              & 1.35              & 1.41              & 2.83              & 2.95              \\
Photo-SLAM~\cite{photoslam}               & 0.39              & 0.41              & \cellcolor{tabthird}0.28              & \cellcolor{tabthird}0.31              & 0.26              & 0.29              & \cellcolor{tabthird}0.27              & \cellcolor{tabthird}0.29              & \cellcolor{tabthird}0.43              & \cellcolor{tabthird}0.45              \\
Loopy-SLAM~\cite{loopyslam}              & 1.31              & 1.47              & 0.93              & 0.99              & 0.74              & 0.69              & 1.15              & 1.03              & 1.95              & 2.01              \\
\multicolumn{11}{l}{\cellcolor[HTML]{EEEEEE}{\textit{Multi-Agent Methods}}} \\
CP-SLAM~\cite{CP-SLAM}                  & 12.89              & 14.56              & 5.63              & 5.98             & 10.01              & 11.63             & 9.47               & 10.49             & 9.50              & 10.67             \\Swarm-SLAM~\cite{swarmslam}                & 0.49              & 0.53              & 0.35              & 0.39              & 0.36              & 0.39              & 0.41              & 0.45            & 0.82              & 0.85              \\
MNE-SLAM~\cite{mneslam}                     & \cellcolor{tabsecond}0.07               & \cellcolor{tabsecond}0.08              & \cellcolor{tabfirst}0.10             & \cellcolor{tabfirst}0.12               & \cellcolor{tabsecond}0.05             & \cellcolor{tabsecond}0.06            & \cellcolor{tabsecond}0.06             & \cellcolor{tabsecond}0.05               & \cellcolor{tabsecond}0.38              & \cellcolor{tabsecond}0.40              \\
Ours                     & \cellcolor{tabfirst}0.06               & \cellcolor{tabfirst}0.07              & \cellcolor{tabfirst}0.10             & \cellcolor{tabfirst}0.12               & \cellcolor{tabfirst}0.04             & \cellcolor{tabfirst}0.05            & \cellcolor{tabfirst}0.05             & \cellcolor{tabfirst}0.05               & \cellcolor{tabfirst}0.35              & \cellcolor{tabfirst}0.38              \\ \bottomrule
\end{tabular}}}
 
\caption{Multi-agent camera tracking performance on DES dataset (Long corridor scene) with sensor depth images. ``-" indicates invalid results due to the failure or inability of these methods. }
  
\end{table}

\noindent\textbf{Trajectory Groundtruth}
Many existing datasets face challenges related to the accuracy of ground truth estimates, shown in Tab.~\ref{tab:datasetcompare}. GNSS/INS is the most commonly used localization method for automotive vehicles, often exhibits errors at the decimeter level. The Motion Capture system is alos primarily effective indoors or in low-light conditions. MoCap
offer accuracy at the centimeter to millimeter level with clear line-of-sight. Two commonly used methods for trajectory estimation through point cloud matching are Normal Distributions Transform (NDT) and Iterative Closest Point (ICP). However, these methods often introduce errors in the range of a few decimeters. Simultaneous Localization and Mapping (SLAM) uses onboard LIDAR sensors but may suffer from long-term drift and is generally considered less accurate, with measurement noise ranging from a few decimeters to several meters. We use a prior map continuous time registration (MCTR) method~\cite{mcd} and CTE-MLO~\cite{ctemlo} to provide centimeter-level accuracy without any line-of-sight requirements, making it a promising choice for creating large datasets with centimeter accuracy. Users can use CEVA (Continuous-time Evaluation)~\cite{mcd} which is a python wrapper for the basalt, library to make working with continuous-time trajectory ground truth easy. In many aspects, CEVA exceeds the basalt library in its utilities.

\begin{table}[!t]
\centering
\scalebox{0.98}{
\setlength{\tabcolsep}{2.5mm}{
\begin{tabular}{lccc}
\toprule
Methods   & Mean [m]$\downarrow$ & Median [m]$\downarrow$ & RMSE [m]$\downarrow$ \\ \midrule
\multicolumn{4}{l}{\cellcolor[HTML]{EEEEEE}{\textit{Single-Agent Methods}}}\\
ORB-SLAM3\cite{orbslam3}  & 3.98     & 3.57       & 4.13    \\
Co-SLAM\cite{coslam}   & -     &  -      & -     \\
ESLAM\cite{eslam}     & 6.54     & 6.27       & 7.13      \\
GO-SLAM\cite{goslam}   & \cellcolor{tabthird}2.59     &  \cellcolor{tabthird}1.98      & \cellcolor{tabthird}3.03     \\
PLGSLAM\cite{plgslam}   & 6.31     & 6.01       & 6.89     \\
PhotoSLAM\cite{photoslam} & 3.58     & 3.17       & 3.27    \\
\multicolumn{4}{l}{\cellcolor[HTML]{EEEEEE}{\textit{Multi-Agent Methods}}} \\
CP-SLAM\cite{CP-SLAM}   & 8.51     & 8.37       & 9.45 \\
MNE-SLAM~\cite{mneslam}  & \cellcolor{tabsecond}2.44    & \cellcolor{tabsecond}1.92       & \cellcolor{tabsecond}2.94  \\
Ours  & \cellcolor{tabfirst}2.38    & \cellcolor{tabfirst}1.86       & \cellcolor{tabfirst}2.89  \\ \bottomrule  
\end{tabular}}}
 
\caption{Camera Tracking Performance on DES dataset (outdoor carpark scene) with sensor depth images. ``-" indicates invalid results due to the failure or inability of these methods. }
  
\label{tab:outdoor}
\end{table}

\begin{figure}[t]
    \centering
    \includegraphics[width=\linewidth]{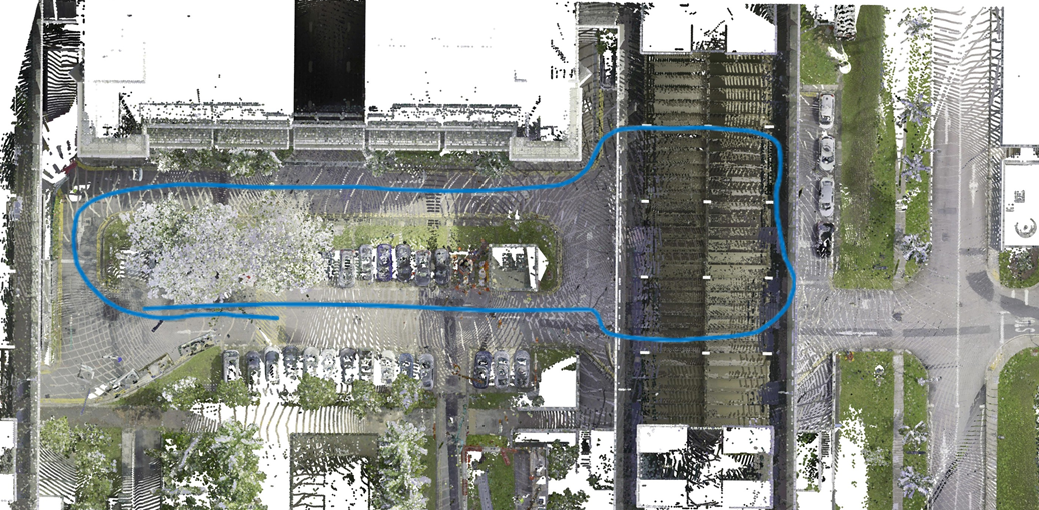}
       
    \caption{
   Outdoor experiments of our proposed framework’s camera tracking performance on the outdoor carpark sequence of DES dataset. }

    \label{fig:outdoor}
\end{figure}

\begin{table*}[]
\centering
\scalebox{0.9}{
\setlength{\tabcolsep}{0.4mm}{
\begin{tabular}{l|lccccccccc}
\toprule
 \multirow{2}{*}{} & \multirow{2}{*}{Methods} & \multirow{2}{*}{Map/Frame} & \multirow{2}{*}{Track/Frame} & \multirow{2}{*}{Memory} & \multicolumn{5}{c}{Communication[MB]} & \multirow{2}{*}{Bandwidth}                                 \\
                         &                          &                          &                            &                         & Visual Descriptor & Network Parameter(Map) & Pose   & Images   & Total &   \\ \midrule
\multirow{6}{*}{Replica\cite{replica}} & ESLAM\cite{eslam}                    & 0.29s                    & 68.5ms                     & 20.73MB                 & -                 & -                & -      & -        &-    &-     \\
                         & Co-SLAM\cite{coslam}                  & 0.15s                    & 54.8ms                     & 26.39MB                 & -                 & -                & -      & -        & -   &-     \\
                         & Point-SLAM\cite{pointslam}               & 10.47s                   & 29.7ms                     & 31.42MB                 & -                 & -                & -      & -        & -    & -    \\
                         & D-GNC\cite{kimera-multi}                  & -                        & -                          & -                       & 2.13MB            & 30.15MB          & 0.45MB & -        & 32.73MB & - \\
                         & CP-SLAM\cite{CP-SLAM}                  & 10.45s                   & 34.7ms                     & 39.75MB                 & 1.57MB            & 39.75MB          & 0.34MB & 388.46MB & 430.38MB & - \\
                    & MNE-SLAM~\cite{mneslam}                     & 0.15s                    & 22.1ms                     & \textbf{24.45MB}                 & 35.12MB            & \textbf{24.45MB}          & 0.34MB & -        & 60.58MB & 0.58MB/s \\
                         & Ours                     & \textbf{0.14s}                    & \textbf{22.1ms}                     & 24.73MB                 & \textbf{0.21MB}            & 24.73MB          & \textbf{0.34MB} & -        & \textbf{25.28MB} & \textbf{0.25MB/s} \\ \midrule
\multirow{6}{*}{Scannet\cite{scannet}} & ESLAM\cite{eslam}                    & 0.70s                    & 0.23s                      & 21.48MB                 & -                 & -                & -      & -        & -   & -     \\
                         & Co-SLAM\cite{coslam}                  & 0.20s                    & 78.3ms                     & 27.19MB                 & -                 & -                & -      & -        & -    & -    \\
                         & Point-SLAM\cite{pointslam}               & 12.69s                   & 64.5ms                     & 45.21MB                 & -                 & -                & -      & -        & -   & -     \\
                         & D-GNC\cite{kimera-multi}                  & -                        & -                          & -                       & 6.37MB            & 38.21MB          & 0.85MB & -        & 45.43MB & - \\
                         & CP-SLAM\cite{CP-SLAM}                  & 10.59s                   & 37.4ms                     & 53.75MB                 & 4.21MB            & 65.75MB          & 0.73MB & 395.18MB  & 465.06MB & - \\
                         & MNE-SLAM~\cite{mneslam}                     & 0.21s                    & 20.8ms                     & \textbf{27.77MB}                 & 38.38MB            & \textbf{27.77MB}          & 0.73MB & -        & 66.88MB & 0.26MB/s \\
                         & Ours                     & \textbf{0.20s}                    & \textbf{20.8ms}                     & 27.89MB                 & \textbf{0.57MB}            & 27.89MB          & \textbf{0.73MB} & -        & \textbf{29.19MB}  & \textbf{0.11MB/s}\\ \bottomrule
\end{tabular}}}
 
\caption{Runtime, Scene representation memory, and multi-robot communication performance evaluation on Replica~\cite{replica} and ScanNet~\cite{scannet} dataset. ``Memory" refers to the storage consumption of the scene representation and the decoder.  ``Communication" refers to the amount of data transmitted during the robot's operation. ``VD" denotes the visual descriptor. ``NP(map)" denotes the network parameter or the sparse point cloud map. }
  
\label{tab:runtime}
\end{table*}

\begin{figure*}[ht]
    \centering
    \includegraphics[width=\linewidth]{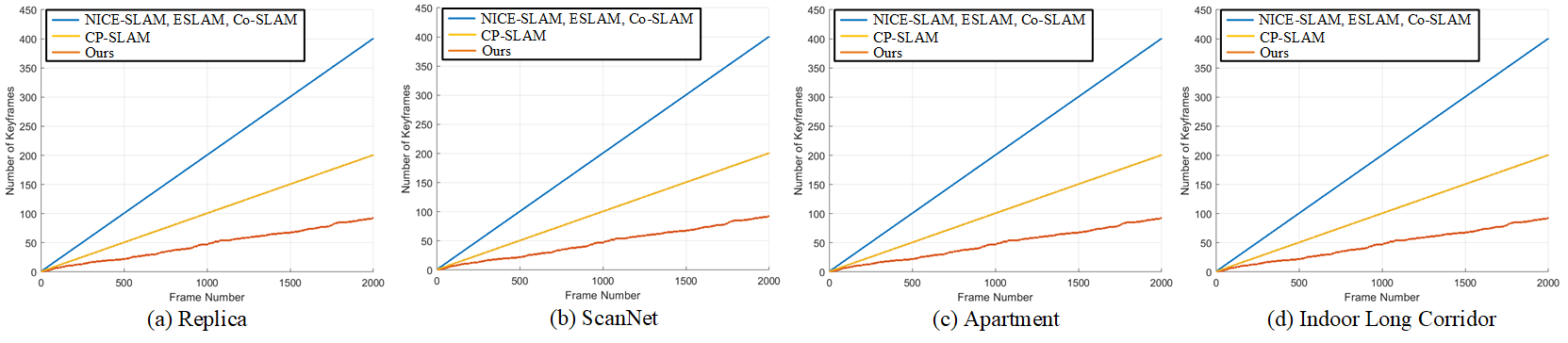}
      
    \caption{
    We compared our method with current state-of-the-art traditional and neural SLAM methods: NICE-SLAM~\cite{niceslam}, Co-SLAM~\cite{coslam}, ESLAM~\cite{eslam}, CP-SLAM~\cite{CP-SLAM} across four different datasets: Replica~\cite{replica}, ScanNet~\cite{scannet}, Apartment~\cite{replica}, and our own dataset. We focus on the number of optimized keyframes. Our keyframe selection approach achieves better reconstruction results while using significantly fewer keyframes than other methods, greatly improving both system efficiency and accuracy. }
    \label{fig:keyframe}
      
\end{figure*}

\noindent \textbf{Comparison with other datasets}
In Tab.~\ref{tab:datasetcompare}, we have reviewed and compared existing datasets from various perspectives, including indoor versus outdoor scenes, sensor modalities (camera, depth, LiDAR, IMU), the availability of ground truth for pose and 3D maps (critical for dense SLAM), the methods used to obtain pose and 3D map ground truth, and a brief description of each dataset. The upper half of the comparison table focuses on 3D reconstruction datasets, while the lower half are SLAM datasets. It is evident that current SLAM datasets do not include 3D map ground truth, while 3D reconstruction datasets are generally unsuitable for SLAM methods. For instance, the ScanNet++ dataset contains non-time-continuous trajectories with numerous abrupt jumps and teleportations, making it incompatible with SLAM systems.

Our dataset is the first Neural SLAM dataset to provide both high-precision 3D map and trajectory ground truth. It features diverse sensor modalities and spans a wide range of scenes, from small indoor rooms to large-scale campus environments, offering a comprehensive large-scale dataset for advancing SLAM research.

\subsection{Experimental Results}

\noindent \textbf{Scene reconstruction }
We evaluate the 3D scene reconstruction performance of our method on the Replica~\cite{replica}, ScanNet~\cite{scannet}, Apartment~\cite{replica}, and our own dataset \textbf{DES}.  As shown in Tab.~\ref{tab:replica}, We demonstrate the effectiveness of our method across eight small room scenes. We present the two agents reconstruction results and the global scene reconstruction performance in small room scenarios. Best results are highlighted as \colorbox{tabfirst}{first}, \colorbox{tabsecond}{second}, and \colorbox{tabthird}{third}. For single-agent methods, we evaluate their performance separately on the sequence of Agent 1, Agent 2, and the full sequence(global). For multi-agent methods, we run the sequences of both Agent 1 and Agent 2 simultaneously. Our  framework outperforms all other single-agent and multi-agent methods, achieving state-of-the-art performance in both single agent reconstruction and global map reconstruction. This success is attributed to our hybrid neural scene representation, which enables accurate and efficient reconstruction. Additionally, the design of distributed camera tracking with intra-to-inter loop  closure and multiple submap fusion ensures consistency across the entire map. In Tab.~\ref{tab:render}, we evaluate the two-agent rendering performance and depth estimation results of our method and other SOTA methods. The results demonstrate that our method achieves high-quality rendering and accurate depth estimation. Our proposed hybrid scene representation effectively capture both high-frequency and low-frequency features of the environment, enabling accurate and complete reconstruction with fine details. In Fig.~\ref{fig:scannet} and Fig.~\ref{fig:apartment}, we present the two-agent surface reconstruction performance on large room scenes from the ScanNet dataset~\cite{scannet} and the Apartment dataset~\cite{replica}. We show the comparison of our method with single-agent methods and multi-agent methods of the global map. The region outlined on the image is marked in red to signify lower reconstruction accuracy, in green to signify higher accuracy, and in yellow to represent the groundtruth results. The experiment results demonstrate the effectiveness of the proposed scene representation method. We show the qualitative comparison on \textbf{DES} dataset of our method with the SOTA method ESLAM in Fig. \ref{fig:indoor} and Fig.~\ref{fig:teaser2}. Our method successfully achieves consistent completion as well as high-fidelity surface reconstruction results in large-scale, real-world indoor scenes. We present the four-agent scene reconstruction results in Tab.~\ref{tab:indoor},  showing that most neural SLAM methods experience significant pose drift in real-world scenarios, ultimately leading to reconstruction failure. To better highlight the advantages of neural SLAM methods, we visualize the rendering performance on the DES dataset in Fig.~\ref{fig:render}, Fig.~\ref{fig:teaser} and quantitative results of rendering performance of training views and novel views in Tab.~\ref{tab:renderview}. Overall, our designed hybrid representation, intra-to-inter loop closure, and multi-submap online distillation, achieves high-precision reconstruction and rendering performance.

\noindent \textbf{Camera tracking}
As shown in Tab.~\ref{tab:replica}, we present the two-agent camera tracking results (Agent1, Agent2) and the global tracking performance in small room scenarios on the Replica dataset. We present the results in terms of RMSE, Mean, and Median. Our method achieves superior tracking accuracy compared to other single-agent and multi-agent approaches. In Tab.~\ref{tab:apartment} and Tab.~\ref{tab:scannet}, we provide two-agent camera tracking performance on Apartment dataset~\cite{replica} and ScanNet dataset~\cite{scannet} for larger indoor scenes. We achieve the SOTA  performance in global camera tracking. This is attributed to our multi-agent collaborative tracking, which leverages intra-to-inter loop closure to eliminate pose error and achieve consistency across multiple agents. Our method outperforms existing multi-agent and single-agent methods. In Tab.~\ref{tab:indoor}, Tab.~\ref{tab:outdoor}, and Fig.~\ref{fig:outdoor},  We present the performance on the DES dataset, showing that most neural SLAM methods experience significant pose drift in real-world scenarios with sensor depth, ultimately leading to reconstruction failure. Our multi-agent approach, however, successfully achieves accurate reconstruction and pose estimation in large-scale environments.

\begin{table}[!t]
\centering
\scalebox{0.9}{
\setlength{\tabcolsep}{0.9mm}{
\begin{tabular}{l|ccc|ccc}
\toprule
\multirow{2}{*}{Methods} & \multicolumn{3}{c|}{Reconstruction{[}cm{]}}       & \multicolumn{3}{c}{Localization{[}cm{]}}         \\
                         & Acc.$\downarrow$           & Comp.$\downarrow$          & Comp.Ratio(\%)$\uparrow$  & Mean           & Median         & RMSE           \\ \midrule
                         w/o tri-plane enc.          & 1.933          & 1.671          & 95.174          & 0.359          & 0.366          & 0.402          \\
                         w/o hashgrid enc.          & 1.739          & 1.581          & 96.034          & 0.359          & 0.366          & 0.402          \\
                         w/o coordinate enc.          & 1.837          & 1.682          & 95.479          & 0.358          & 0.366          & 0.402          \\
w/o intra-loop.          & 1.831          & 1.592          & 95.779          & 0.397          & 0.403          & 0.432          \\
w/o inter-loop           & 1.903          & 1.597          & 95.679          & 0.401          & 0.409          & 0.441          \\
w/o submap               & 1.949          & 1.607          & 95.494          & 0.359          & 0.365          & 0.401          \\
Ours                     & \textbf{1.679} & \textbf{1.519} & \textbf{96.969} & \textbf{0.355} & \textbf{0.360} & \textbf{0.394} \\ \bottomrule
\end{tabular}}}
 
\caption{ Ablation study. We conduct experiments on Replica dataset~\cite{replica} to verify the effectiveness of our method. Our full model achieves better
completion reconstructions and more accurate pose estimation results.}
  
\label{tab:ablation}
\end{table}

\noindent\textbf{Time analysis}
We analyze the speed and memory usage of our method compared with other SOTA methods in Replica datasets~\cite{replica} and ScanNet datasets~\cite{scannet}. We evaluate the time consumption of tracking and mapping every iteration and every frame in Tab.~\ref{tab:runtime}. We also evaluate the memory and communication consumption of existing methods. We provide a detailed breakdown of the data types transmitted during inter-robot communication. Our method transmits visual descriptors, keyframe poses, and neural network parameters. In contrast, centralized methods such as CP-SLAM also require transmitting full keyframe images. Sparse point cloud-based methods like Kimera-Multi~\cite{kimera-multi} typically transmit sparse point cloud maps, feature descriptors, and data for distributed pose graph optimization. Note that Kimera-Multi requires IMU data to run its odometry module. Therefore, in our evaluation, we only validate the communication process and replace the original odometry with ORB-SLAM for comparison. Compared to CP-SLAM, our method improves communication efficiency by seven times (\textbf{7} $\times$) with only bandwidth-efficient information is exchanged, including neural network parameters, keyframe descriptors, and pose. Our distributed method significantly enhances communication efficiency, increasing the effectiveness in real-world operations. In Fig.~\ref{fig:keyframe}, we compare our method with the current SOTA neural SLAM methods of keyframe selection strategy. Our keyframe selection approach achieves better reconstruction results
while using significantly fewer keyframes than other methods, greatly improving both system efficiency and accuracy.

\subsection{Ablation Study}
In this section, we conduct various experiments to verify the effectiveness of our method. 
Tab. \ref{tab:ablation} illustrates a quantitative evaluation with different settings.

\noindent\textbf{Hybrid Scene Representation}
We separately remove our triplane scene representation and coordinate encoding in these experiments.  Our full model leads to higher scene reconstruction accuracy and better completion. The triplane scene representation is well-suited for modeling low-frequency signals with fast convergence speed, while the hashgrid network excels at capturing high-frequency details. This enables high-precision and efficient modeling of scene information. The coordinate encoding network improves the reconstruction completeness and enables certain levels of hole filling.

\noindent\textbf{Intra-Loop Closure}
We remove our intra loop closure in this experiment. Our full model leads to higher scene reconstruction accuracy and better completion. The intra loop closure can significantly reduce the growing cumulative error and improve the robustness and accuracy of the single-agent camera tracking.

\noindent\textbf{Inter-Loop Closure}
We remove the inter-loop closure in this experiment. Inter-loop closure can effectively register different submaps into the global map and improve the tracking performance across different agents.

\noindent\textbf{Online Distillation}
Submap fusion can significantly improve the integration of information among different maps, ensuring the consistency of the global map; otherwise, the map may exhibit numerous discontinuities or floating artifacts. This also suggests that handling smooth transitions between two submaps, especially when revisiting an inactive submap, is critical to the overall reconstruction quality.

\section{Conclusion}
In this paper, we propose a novel distributed multi-agent neural SLAM system, MCN-SLAM, which achieves accurate collaborative surface reconstruction and pose estimation in both small and large-scale indoor scenes. Our distributed mapping and tracking framework enables our system to represent large-scale indoor scenes under communication bandwidth restriction. The intra-to-inter loop closure can effectively eliminate pose drifts and achieve global map consistency across multiple agents. Our extensive experiments demonstrate the effectiveness and accuracy of our system in both scene reconstruction and pose estimation. The proposed neural SLAM dataset with high-accuracy 3D mesh and time-continuous camera trajectory can greatly advance the development of the community.

\bibliographystyle{IEEEtran}
\bibliography{egbib}

\end{document}